\documentclass[10pt,twocolumn,letterpaper]{article}

\usepackage{cvpr}              

\definecolor{cvprblue}{rgb}{0.21,0.49,0.74}
\usepackage[pagebackref,breaklinks,colorlinks,allcolors=cvprblue]{hyperref}
\usepackage[accsupp]{axessibility}  

\def\paperID{20425} 
\def\confName{CVPR}
\def\confYear{2026}

\definecolor{myPurple}{rgb}{0.4, .0, .8}
\definecolor{myGreen}{rgb}{0, .8, .3}
\definecolor{myRed}{rgb}{0.8, .2, .2}
\definecolor{myOrange}{rgb}{0.7, 0.45, 0.2}
\definecolor{myBlue}{rgb}{.0, .0, 1.0}
\definecolor{myBlue2}{rgb}{.0, .0, 0.5}
\definecolor{myBlack}{rgb}{.0, .0, 0.0}
\definecolor{red}{rgb}{0.8, 0.2, 0.2}
\usepackage{soul}

\title{PoseMaster: A Unified 3D Native Framework for Stylized Pose Generation}

\author{%
Hongyu Yan$^{1,2}$\footnotemark[1] \quad 
Kunming Luo$^{1}$\footnotemark[1] \quad
Weiyu Li$^{1}$ \quad 
Kaiyi Zhang$^{1,2}$ \quad \\
Yixun Liang$^{1}$ \quad
Jingwei Huang$^{2}$ \quad 
Chunchao Guo$^{2}$\footnotemark[2] \quad 
Ping Tan$^{1}$\footnotemark[2]
\\ \vspace{-10pt}\\
$^{1}$Hong Kong University of Science and Technology \qquad 
$^{2}$Tencent Hunyuan \qquad  \\
\{hyanar, kluoad\}@connect.ust.hk, pingtan@ust.hk
}

\begin{document}
\twocolumn[{%
\renewcommand\twocolumn[1][]{#1}%
\maketitle
\begin{center}
    \centering
    \captionsetup{type=figure}
    \includegraphics[width=1.0\linewidth]{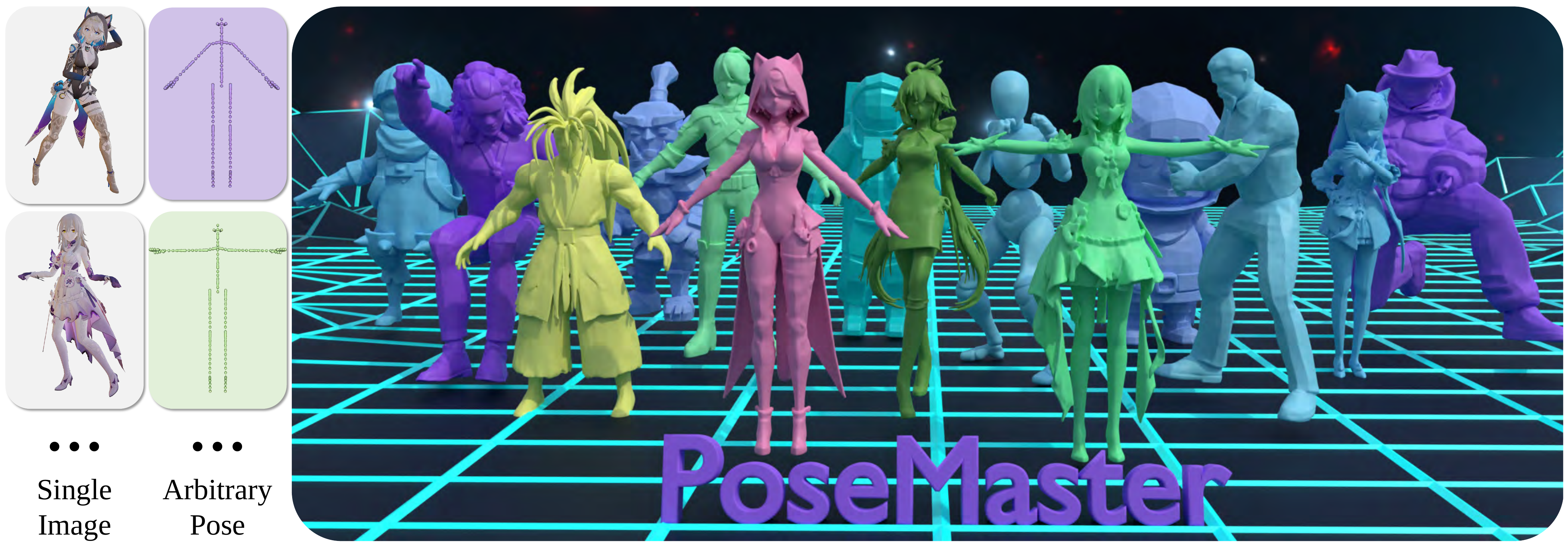}
    \vspace{-0.1in}
    \captionof{figure}{Given a single image and arbitrary poses represented by a 3D skeleton, PoseMaster can generate a high-quality 3D asset that maintains the identity of the image while adhering to the pose defined by the skeleton, enabling rich and precise 3D  pose stylization.}
    \label{fig:teaser}
\end{center}%
}]
\maketitle

\renewcommand{\thefootnote}{\fnsymbol{footnote}} 
\footnotetext[1]{Equal contribution.} 
\footnotetext[2]{Corresponding authors.}

\begin{abstract}
Pose stylization, which aims to synthesize stylized content aligning with target poses, serves as a fundamental task across 2D, 3D, and video domains. In the 3D realm, prevailing approaches typically rely on a cascade pipeline: first manipulating the image pose via 2D foundation models and subsequently lifting it into 3D representations. However, this paradigm limits the precision and diversity of the 3d pose stylization. To this end, we propose a novel paradigm for 3D pose stylization that unifies pose stylization and 3D generation within a cohesive framework. This integration minimizes the risk of cumulative errors and enhances the model’s efficiency and effectiveness. In addition, diverging from previous works that typically utilize 2D skeleton images as guidance,  we directly utilize the 3D skeleton because it can provide a more accurate representation of 3D spatial and topological relationships, which significantly enhances the model’s capacity to achieve richer and more precise pose stylization. Moreover, we develop a scalable data engine to construct a large-scale dataset of ``Image-Skeleton-Mesh" triplets, enabling the model to jointly learn identity preservation and geometric alignment. Extensive experiments demonstrate that PoseMaster significantly outperforms state-of-the-art methods in both qualitative and quantitative metrics. Owing to the strict spatial alignment between the generated 3D meshes and the conditioning skeletons, PoseMaster enables the direct creation of animatable assets when coupled with automated skinning models, highlighting its compelling potential for automated character rigging.

\end{abstract}    
\section{Introduction}
\label{sec:intro}
With the burgeoning demand for digital content in gaming, film production, and virtual reality, the ability to generate stylized assets that maintain unified Intellectual Property (IP) consistency has become paramount. While significant strides have been made in 2D and video generation, 3D pose stylization—the task of generating 3D assets that strictly align with a target pose while preserving specific visual styles—remains a formidable challenge. Although recent 3D native generative models have demonstrated the capacity to synthesize high-quality geometry from a single image, controlling these generations with precise pose guidance continues to be an underexplored frontier.

Prevailing approaches typically treat 3D pose stylization as a decoupled process, combining 2D pose manipulation with subsequent 3D lifting. Methods such as CharacterGen~\cite{peng2024charactergen}, StdGen~\cite{he2024stdgen}, and SKDream~\cite{xu2025skdream} rely heavily on a cascade pipeline: they first employ 2D controllable models (e.g., ControlNet~\cite{zhang2023adding}) to generate a stylized image guided by a 2D skeleton image, and then utilize a separate reconstruction model to lift this image into a 3D asset. Despite their popularity, these cascade paradigms suffer from two fundamental limitations. First, error propagation is inevitable; artifacts, occlusions, or inconsistencies introduced during the 2D generation phase are directly amplified in the 3D reconstruction, leading to geometric distortions. Second, relying on 2D skeleton images introduces inherent geometric ambiguity. A 2D projection loses critical depth information and spatial relationships, making it difficult to resolve self-occlusions or complex topological structures, thereby limiting the precision of the final 3D pose.

To bridge this gap, we present PoseMaster, a novel 3D native framework that unifies pose stylization and 3D generation within a single, end-to-end generative model. Diverging from the conventional reliance on 2D skeleton images, we propose to explicitly leverage 3D skeletons as the control signal. Unlike 2D representations which suffer from projection ambiguity, 3D skeletons provide explicit spatial coordinates and topological relationships, offering a robust geometric prior for the generation process. To harness this, we design a specialized 3D skeleton encoder capable of capturing fine-grained structural details. This design enables PoseMaster to bypass the limitations of the 2D-lifting paradigm, minimizing cumulative errors and ensuring that the generated 3D assets are precisely aligned with the target pose while maintaining high-fidelity identity preservation.

However, training such a 3D native model presents a significant data challenge: there is a scarcity of large-scale, high-quality paired data containing ``Image-Skeleton-3D Mesh" triplets. Naive solutions, such as utilizing existing animatable datasets like ReadyPlayerMe~\cite{Readyplayerme} or VRoid~\cite{vroid}, often fall short in terms of stylistic diversity and quantity. To address this bottleneck, we introduce a scalable Data Engine designed to construct a massive dataset of stylized 3D assets by introducing both static and animated 3D assets, which enables the model to jointly learn the correlation between visual appearance (ID) and geometric structure (Pose), significantly enhancing its generalization capabilities across diverse styles and complex poses.

Extensive evaluations demonstrate that PoseMaster establishes a new state-of-the-art in both pose canonicalization and arbitrary-pose stylization. Crucially, the 3D meshes generated by our framework exhibit precise spatial alignment with the conditioning 3D skeletons. This inherent spatial consistency naturally yields rigged assets that can be seamlessly skinned and animated, entirely bypassing the need for time-consuming manual skeleton retargeting and post-processing.


\begin{itemize}
\item We propose PoseMaster, a novel 3D native generation paradigm for pose stylization, which integrates pose stylization and 3D geometry generation into a cohesive framework, effectively eliminating the error propagation inherent in traditional cascade pipelines.

\item We introduce a novel 3D-skeleton-based control mechanism and a scalable data engine. By capturing explicit spatial and topological relationships from 3D skeletons and training on large-scale ``Image-Skeleton-Mesh" triplets, we significantly enhance the model's precision and generalization ability.

\item We demonstrate that our method achieves state-of-the-art performance in both qualitative and quantitative evaluations. Furthermore, we validate the practical value of our framework by showing that it generates rigged assets to facilitate the animatable character creation.

\end{itemize}

\begin{figure*}[t]
\includegraphics[width=0.95\linewidth]{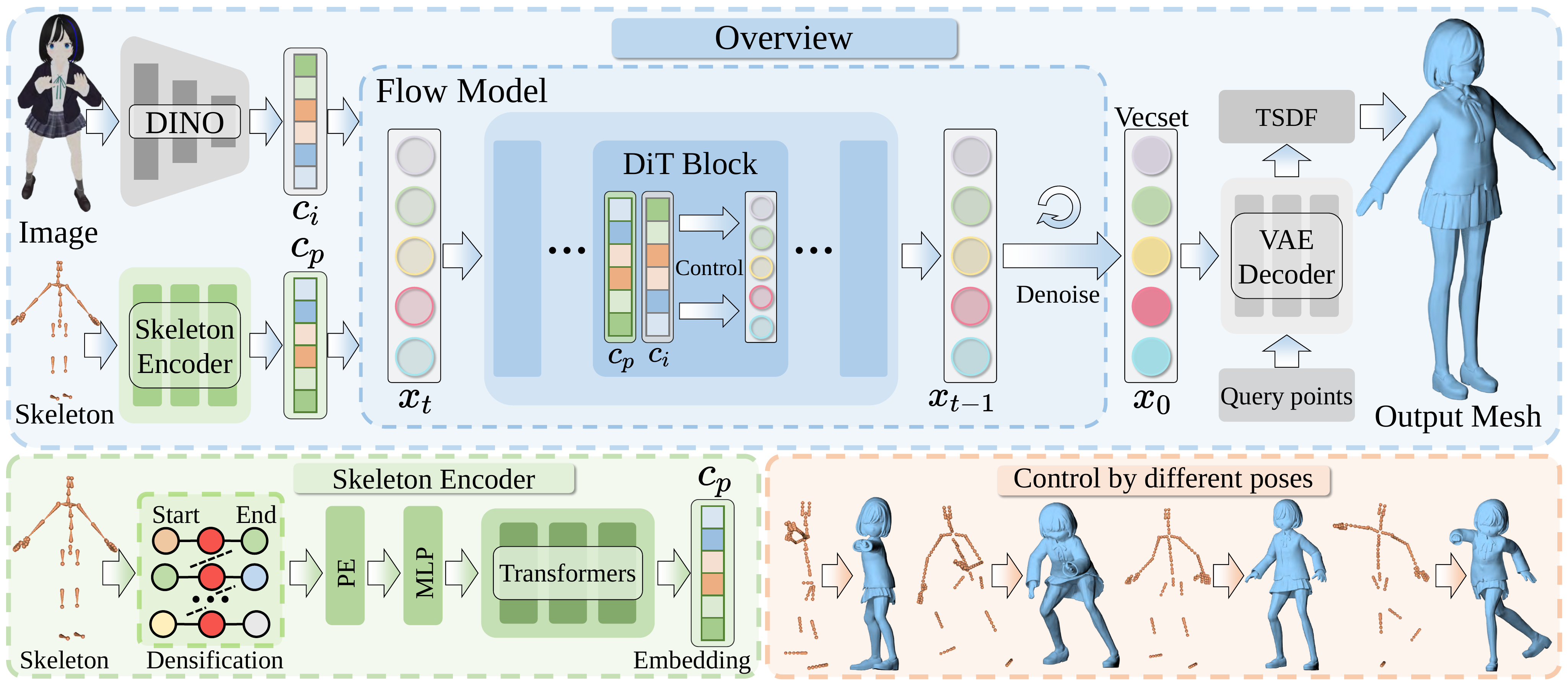}
    \centering
    \caption{Overall pipeline of our PoseMaster. Given a single image and a 3D skeleton, our PoseMaster integrates them into a unified 3D native generation framework to achieve precise pose stylization under 3D generation.}
    \label{fig:overall_pipeline}
\end{figure*}

\section{Related work}

\subsection{2D Pose Stylization}
Pose stylization is a longstanding research challenge in the field of image and video generation. Recent advances, such as image diffusion models like Stable Diffusion~\cite{22stablediffusion}, have inspired methods that incorporate a 2D skeleton condition to achieve pose stylization. Notable examples include ControlNet~\cite{zhang2023adding} and its variants~\cite{zhang2025easycontrol,peng2024controlnext,xie2023omnicontrol,zhao2023uni}, which utilize skeletons derived from Openpose~\cite{cao2019Openpose} for pose control. However, these methods are limited to constraining the content generated by text-to-image (T2I) models and do not allow for editing poses in existing images. While Vision-Language Models such as Qwen-Image~\cite{wu2025qwen}, GPT-4o~\cite{GPT4o}, NanoBanala~\cite{Nano}, and Seed~\cite{seed} can synthesize and edit poses from text or image prompts, achieving rich pose stylization while preserving identity remains challenging. In the domain of video generation, several methods~\cite{hu2024animate,wang2024unianimate,tan2024animate,xu2025hypermotion} employ skeleton sequences to create animation videos. Inspired by these, this paper aims to advance 3D pose stylization in the 3D generation framework.

\subsection{3D Pose Stylization}
Early 3D pose stylization approaches~\cite{cao2023dreamavatar, jiang2023avatarcraft, kolotouros2023dreamhuman, liao2024tada, Zhang_Chen_Yang_Qu_Wang_Chen_Long_Zhu_Du_Zheng_2023, zhang2025avatarstudio, zhang2023adding} often rely on 2D stable diffusion priors and human geometry priors from models like SMPL or Openpose~\cite{cao2019Openpose} to constrain generated 3D humans. While these text-based generation models exhibit impressive controllability, they strictly rely on textual descriptions for character specification, thereby lacking support for more intuitive visual or spatial input modalities. To address this, CharacterGen~\cite{peng2024charactergen} introduced an image-driven pipeline for controllable character generation by converting arbitrary-pose images into A-pose multi-view images using multi-view diffusion models~\cite{long2023wonder3d,li2024era3d,luoctr3d}, followed by reconstruction of A-pose 3D characters through a multi-view-based LRM~\cite{hong2023lrm}. Building on CharacterGen, StdGen~\cite{he2024stdgen} and SKDream~\cite{xu2025skdream} enhance geometric quality and controllability. However, these approaches depend on 2D models for pose stylization, which can lead to identity and texture distortion and cumulative errors. In this paper, we propose a native paradigm for cohesive pose stylization and 3D generation, achieving superior geometric quality and control accuracy.

\subsection{3D Native Generation}

3D native generation methods utilize representations such as point clouds~\cite{li2018point,zhou20213d,pointflow}, meshes~\cite{nash2020polygen,Liu2023MeshDiffusion,siddiqui2024meshgpt}, and implicit functions~\cite{chen2019learning,park2019deepsdf,ren2024xcube} to optimize their networks. These methods benefit from direct 3D data optimization, resulting in impressive geometric quality. The introduction of the VecSet representation by 3DShapeVecSet~\cite{zhang20233dshape2vecset} has spurred rapid advancements in 3D native generation. Building on VecSet, models such as Michelangelo~\cite{zhao2023michelangelo}, Clay~\cite{zhang2024clay}, CraftsMan~\cite{li2024craftsman}, and Hunyuan3D 2.1~\cite{Hunyuan3D2025Hunyuan3D21} have developed large 3D latent diffusion models to enhance the effectiveness and generalization of 3D generation. Recent methods~\cite{chen2025ultra3d,direct3ds2,triposg,Hunyuan3D2025Hunyuan3D21,lan2024ln3diff,GaussianAnything,3DTopia-XL,wu2024direct3d,stepfun3d,anonymous2025voxset,ye2025hi3dgen} demonstrate impressive geometric quality using sparse voxel representations introduced by Trellis~\cite{xiang2025trellis}. However, these methods often rely heavily on image alignment, limiting their ability to enrich input data. In this paper, we advance the 3D native generation pipeline by integrating 3D skeleton conditions to achieve pose stylization while preserving the identity of the input.

\section{Preliminary Study}
A 3D native generation model comprises two main components: the 3D Variational Autoencoder (VAE) and the 3D Diffusion Model (DiT). The 3D VAE is employed to obtain the latent representation of geometry. For instance, in a VecSet-based VAE, given a surface input, the model first utilizes an encoder to derive the latent VecSet. Subsequently, it recovers the truncated signed distance function (TSDF) through a decoder, allowing us to leverage iso-surface extraction to obtain explicit mesh output.


Flow-based DiT~\cite{liu2022flow} defines the forward process, ${x}_t=(1-t) {x}_1+t {x_0}$, as straight paths between the data distribution and a standard normal distribution. In this design, the velocity $v$ can be approximated with a neural network $v_{\theta}$ by minimizing the conditional flow matching objective:
\begin{equation}
\mathbb{E}_{t, {x}_0, {x_1}}\left\|{v}_\theta({x}, t)-\left({x_1}-{x}_0\right)\right\|_2^2 .
\end{equation}

Following this, we can obtain our conditional flow matching objective:
\begin{equation}
\mathbb{E}_{t, {x}_0, {x_1}, c_{i}, c_{p}}\left\|{v}_\theta({x}, t, c_{i}, c_{p})-\left({x_1}-{x}_0\right)\right\|_2^2 .
\end{equation}
where $c_{i}$ is image condition while $c_{p}$ is pose condition.

\section{Method}
In this section, we begin by introducing our data construction to show how we organize our data to achieve native stylized pose generation in Section~\ref{sec:data}. Then, we present how we design our unified framework to accomplish high-quality pose stylization in Section~\ref{sec:method}.

\begin{figure}
    \centering
    \includegraphics[width=\linewidth]{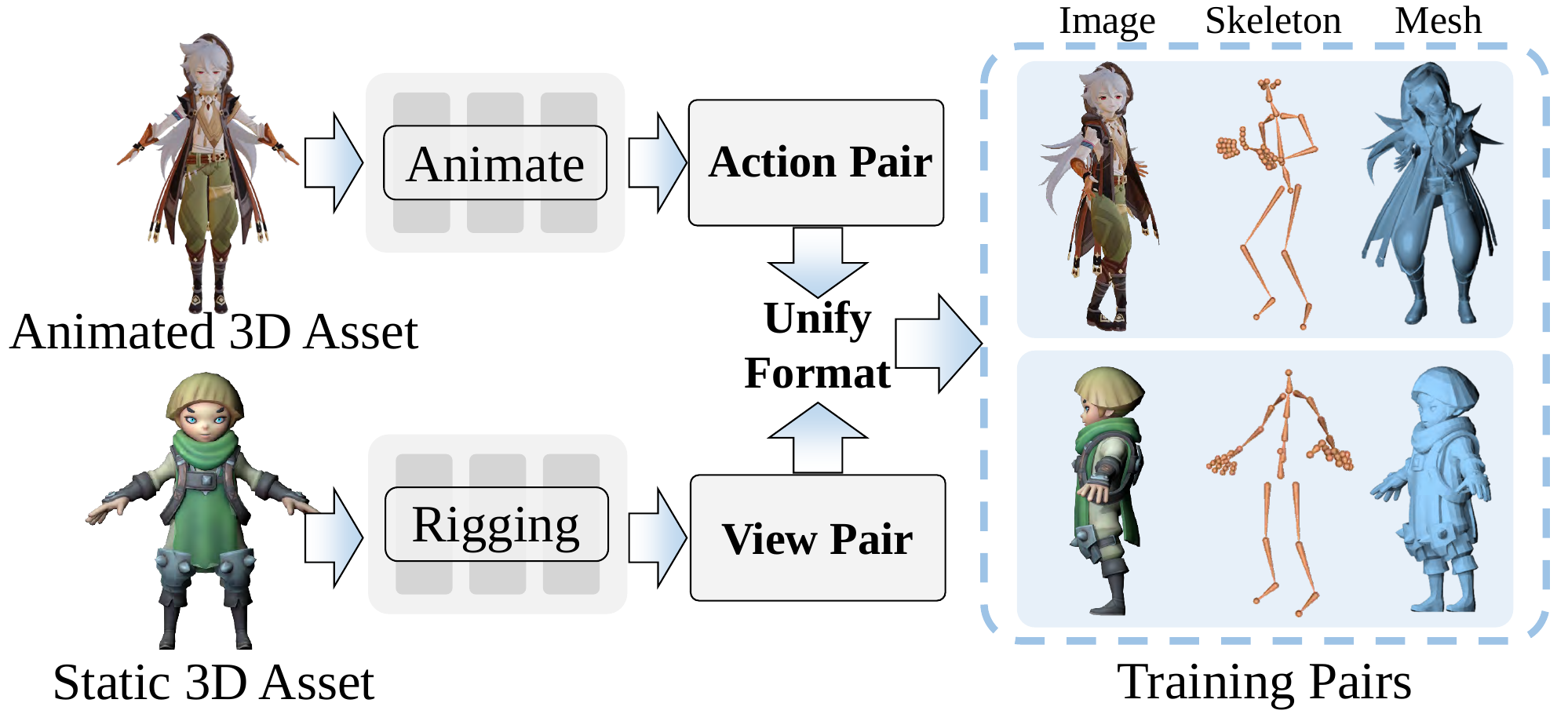}
    \caption{Overview of the dataset construction pipeline. Our approach integrates both dynamic articulated meshes and static geometries to curate a large-scale dataset consisting of strictly aligned image-skeleton-mesh triplets.}
    \label{fig:rendering}
\end{figure}
\subsection{Data Engine}
\label{sec:data}
Achieving end-to-end pose stylization within a native 3D generation framework requires a dataset of cross-pose ``Image-Skeleton-Mesh" triplets, where the 3D structures (skeleton and mesh) are perfectly aligned but strictly decoupled from the reference image pose. While current methods~\cite{peng2024charactergen,he2024stdgen} construct such triplets using animatable 3D assets, this paradigm suffers from two major limitations. First, the severe scarcity of open-source animatable assets hinders the construction of datasets at scale. Second, the limited stylistic diversity of existing animatable assets restricts the model's generalization capabilities. To overcome these bottlenecks, we introduce a scalable data engine.

Specifically, we systematically collect diverse open-source 3D datasets and rigorously filter them to extract humanoid assets. As illustrated in Figure~\ref{fig:rendering}, we synergize both animatable and static 3D assets to synthesize large-scale, pose-decoupled ``Image-Skeleton-Mesh" triplets.

{\bf For animatable 3D assets} (e.g., ReadyPlayerMe~\cite{Readyplayerme}, VRoid~\cite{vroid}, and Playbox~\cite{playbox}), we animate the characters using various motion sequences to render multi-frame images, skeletons, and meshes. To construct cross-pose pairs (denoted as Action Pairs), we sample an image from one motion frame and pair it with the spatially aligned skeleton and mesh extracted from another frame (see Figure~\ref{fig:rendering}).

{\bf For static 3D meshes} sourced from diverse platforms like Objaverse~\cite{deitke2023objaverse}, Objaverse-XL~\cite{deitke2023objaversexl}, and HumanRig~\cite{chu2025humanrig}, we first render multi-view images. To obtain the corresponding 3D skeletons, we adopt a dual strategy: for inherently rigged assets, we directly extract their provided skeletons, whereas for unrigged meshes, we employ an automatic rigging model to infer their skeletal structures. Subsequently, we pair a rendered image from one specific viewpoint with the fully aligned 3D skeleton and mesh to construct View Pairs, which implicitly forces the network to infer view-occluded regions.

Furthermore, we employ the watertight algorithm from Hunyuan3D~2.1 to generate watertight meshes and sample surface points. Crucially, we track the exact normalization parameters and apply identical transformations to the corresponding skeletons, thereby ensuring strict spatial registration between the geometries and their kinematic structures. Ultimately, our data engine yields a massive dataset comprising over 500K unique humanoid objects. By permuting the diverse poses and viewpoints for each object, the total volume of valid training pairs scales well into the millions.

\subsection{Framework}
\label{sec:method}
Similar to recent 3D native generation models~\cite{li2024craftsman,zhang2024clay}, our PoseMaster consists of two primary components: a 3D Variational Autoencoder (VAE) and a 3D Diffusion Transformer (DiT), as illustrated in Figure~\ref{fig:overall_pipeline}. The backbone of our architecture is built upon Hunyuan3D 2.1~\cite{Hunyuan3D2025Hunyuan3D21}, into which we introduce a dedicated skeleton encoder to capture geometric pose constraints. Consequently, given an arbitrary reference image and a target 3D skeleton, our framework generates a high-fidelity 3D mesh that preserves the identity and appearance of the input image while strictly adhering to the pose defined by the skeleton.

\begin{figure}
    \centering
    \includegraphics[width=0.9\linewidth]{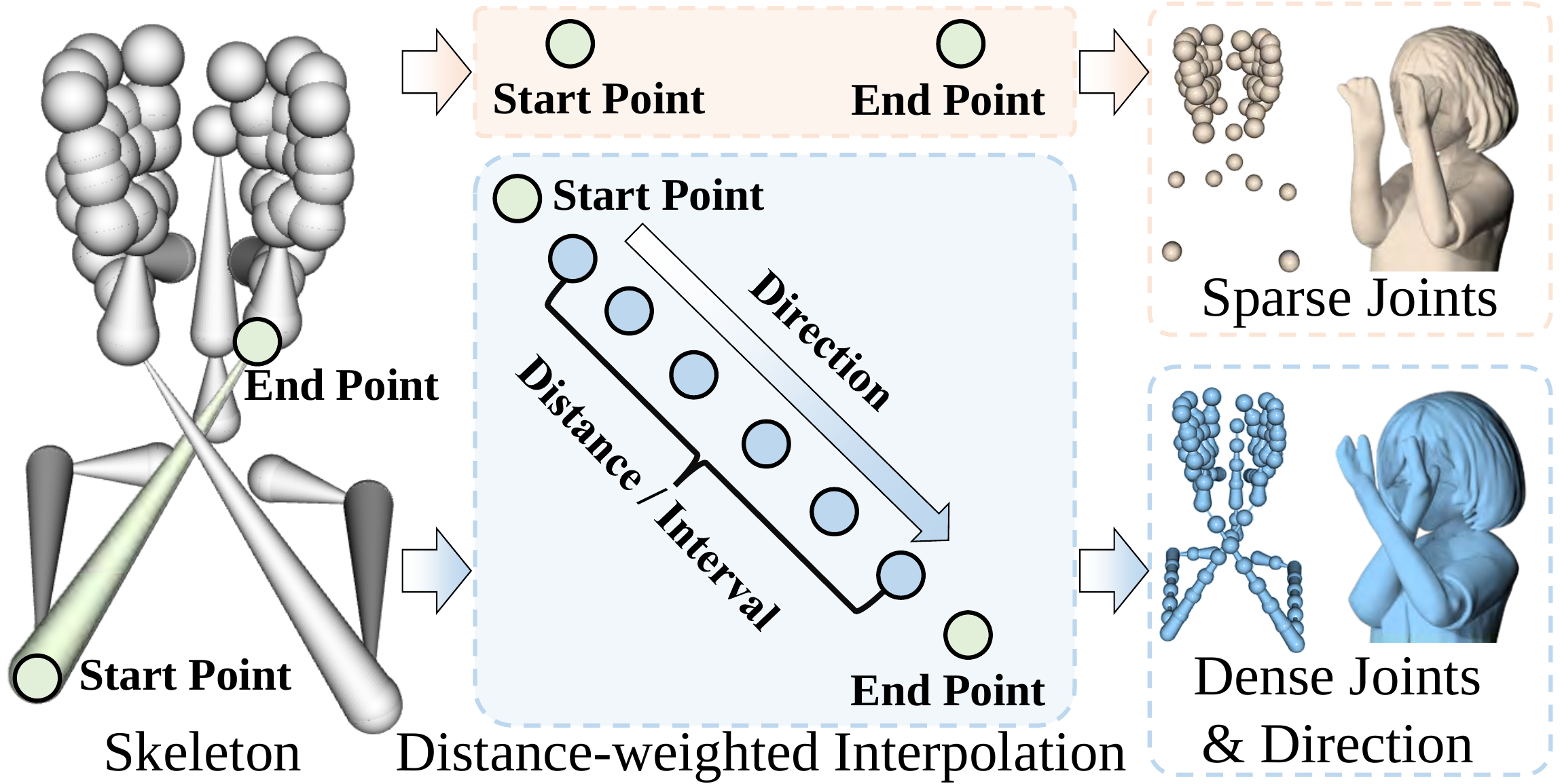}
    \caption{The skeleton representation of sparse joints and ours. We propose a distance-weighted interpolation scheme to densify the skeletal graph. Furthermore, explicit bone directional vectors are embedded into all interpolated points belonging to the same skeletal segment.}
    \label{fig:skeleton-interpolation}
\end{figure}

\subsubsection{Skeleton Encoders}
The skeleton encoder plays a pivotal role in our pipeline. Previous 2D- or video-based pose stylization methods typically utilize 2D skeleton maps (e.g., Openpose~\cite{cao2019Openpose}), which is a natural choice since their output modalities align with the 2D domain. However, for 3D native generation, relying on 2D skeletons fails to provide the network with sufficient spatial and topological awareness, inevitably leading to imprecise control. In contrast, a 3D skeleton offers an explicit geometric scaffold, significantly alleviating the complexity of geometric reconstruction in 3D generation. Thus, employing 3D skeletons as pose conditions is inherently better suited for 3D pose stylization.

A standard 3D skeleton consists of the start and end joints of individual bones. A straightforward approach of representing the pose, as seen in previous work~\cite{zhang2024joint2human}, is the 3D coordinates of sparse joints. While this adequately conveys simple poses (e.g., A-pose or T-pose), its sparsity and lack of explicit topological connectivity hinder the network's ability to interpret complex articulations, as demonstrated in Figure~\ref{fig:skeleton-interpolation}. To address this, we propose a densification strategy to extract structural pose representations. Specifically, we perform distance-weighted interpolation along each bone—from the start joint to the end joint—to construct a dense point cloud. To ensure uniform point distribution regardless of varying bone lengths, the number of interpolated points per bone is strictly determined by its bone length and predefined spatial interval (e.g., 0.005). Furthermore, to explicitly encode topological information into this dense representation, we assign the directional vector of each bone (defined as the vector from the start to the end joint) to all sampled points belonging to that bone. Consequently, our pose representation is formulated as $P \in \mathbb{R}^{N \times 6}$, where $N$ denotes the number of points, and each point contains both 3D coordinates and 3D directional features. Finally, we employ Farthest Point Sampling (FPS) to uniformly downsample the interpolated point cloud to a fixed size of 256 points.

To extract skeleton features, as depicted in Figure~\ref{fig:overall_pipeline}, we leverage a point transformer architecture to capture fine-grained spatial structures and topological relationships. Specifically, the 3D coordinates $P_c$ are first projected into high-dimensional embeddings via a positional encoding layer, and then concatenated with the directional features $P_f$. A linear projection layer subsequently maps the concatenated features to a dimension of 1024, followed by two stacked point transformer blocks to derive the final skeleton condition $c_{p}$. This process is formulated as:
\begin{equation}
c_{p} = \phi_2(\phi_1(\mathcal{T}([PE(P_c), P_f])))
\end{equation}
where $[\cdot]$ denotes concatenation along the feature dimension, $\text{PE}(\cdot)$ represents the positional embedding, $\mathcal{T}(\cdot)$ is the linear projection layer, and $\phi_1, \phi_2$ correspond to the two point transformer blocks.

\subsubsection{Multi-Condition DiT}
Our PoseMaster integrates two condition encoders to process the image condition $c_i$ and the skeleton condition $c_{p}$. For the image input $I_{i}$, following Hunyuan3D 2.1~\cite{Hunyuan3D2025Hunyuan3D21}, we utilize DINOv2-Large~\cite{oquab2023dinov2} at a resolution of 518 to extract the image features, formulated as $c_{i} = \text{DINOv2}(I_{i})$. To preserve the generative priors of the foundation 3D model while allowing the skeleton condition to exert dominant control over the generated pose during training progressively, we integrate the skeleton condition $c_{p}$ with the image condition $c_i$ at the token level to form a joint condition representation. The CFG formulation for predicting the velocity/noise $v_\theta$ is defined as follows:
\begin{equation}
\hat{v}_\theta = v_\theta(x_t, t, c_{p}, \emptyset) + \lambda \cdot \left( v_\theta(x_t, t, c_{p}, c_i) - v_\theta(x_t, t, c_{p}, \emptyset) \right)
\label{eq:freeze}
\end{equation}
where $x_t$ is the noisy input at timestep $t$, $\emptyset$ denotes the nullified image condition, and $\lambda$ is the guidance scale.
\section{Experiments}

\begin{figure*}[t]
\includegraphics[width=1\linewidth]{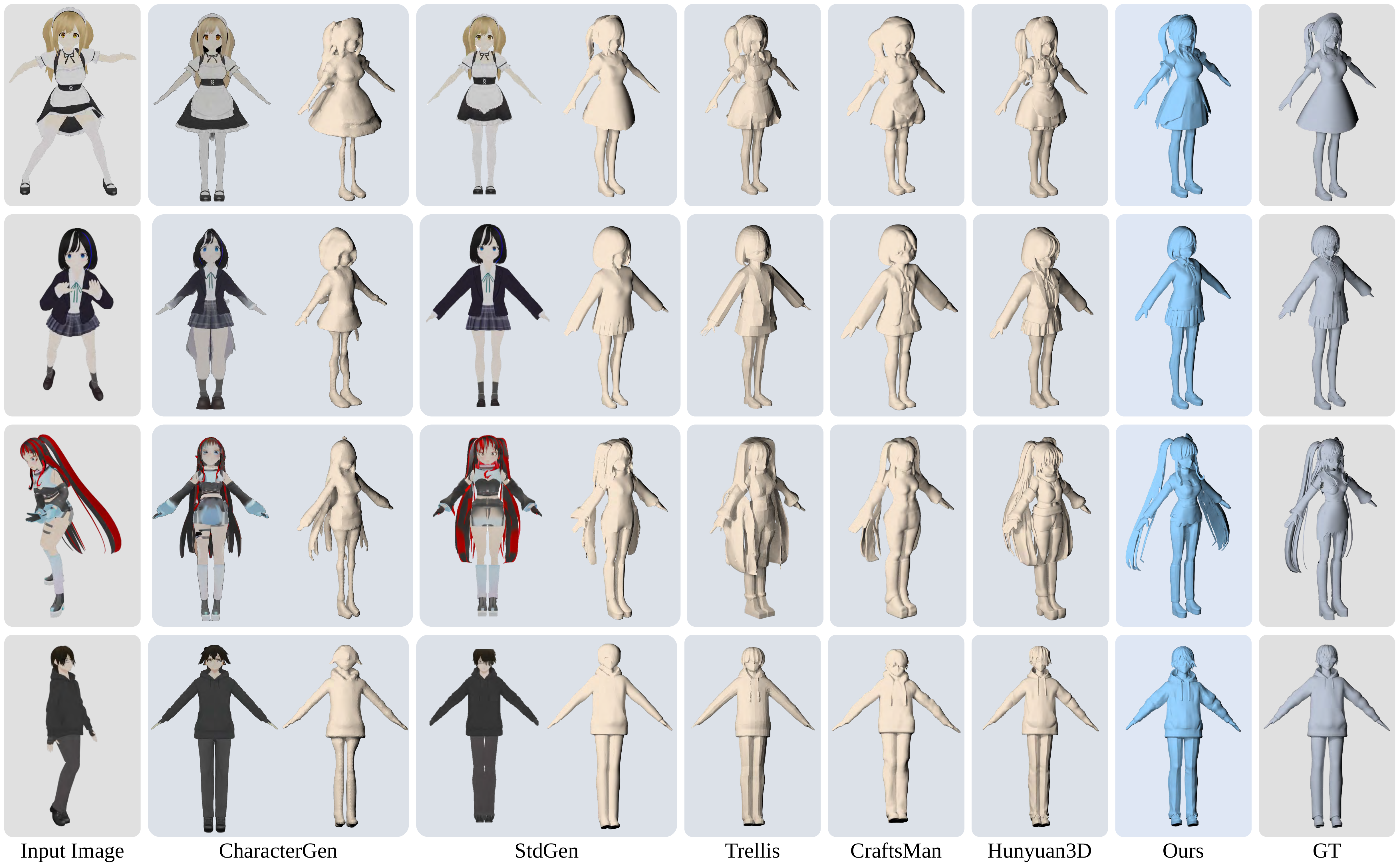}
    \centering
    \caption{The qualitative comparison for pose canonicalization on VRoid~\cite{vroid} dataset. Our method achieves higher geometry quality and accuracy compared with baselines.}
    \label{fig:apose}
\end{figure*}

\subsection{Implementation Details and Metrics}

We initialize our model using the pre-trained weights of the VAE and Diffusion Transformer (DiT) from Hunyuan3D 2.1~\cite{Hunyuan3D2025Hunyuan3D21}. During the fine-tuning phase, we keep the image encoder and VAE strictly frozen. We specifically focus on jointly optimizing the DiT backbone and the skeleton encoder, employing a constant learning rate of $1 \times 10^{-5}$. Furthermore, to ensure dimensional compatibility with the extracted DINO features, the skeleton conditions are projected into a 1024-dimensional feature space. To enhance the robustness and generalization of PoseMaster, we incorporate comprehensive data augmentation strategies across both 2D and 3D modalities. For the 2D image input, we apply random planar rotations $\in[-30^\circ, 30^\circ]$ with a 15\% probability. For the 3D skeleton, we adopt standard point cloud augmentation techniques~\cite{qi2017pointnet++, yan2022fbnet, yan2025symmcompletion}, encompassing random translation, scaling, and rotation. Crucially, to preserve strict spatial alignment between the driving skeleton and the target 3D mesh—a prerequisite for extracting accurate and consistent latent codes—we apply these identical 3D transformations synchronously to both the skeleton and the sampled surface points.


Regarding evaluation metrics, conventional geometric distances such as Chamfer Distance (CD) are sub-optimal for assessing 3D pose stylization. This limitation arises because generative models inherently struggle to achieve perfect spatial alignment with ground-truth (GT) meshes, particularly in high-variance (e.g., garment, hair) or occluded regions. Therefore, following recent robust surface evaluation protocols~\cite{fu2024geowizard,ke2025marigold}, we evaluate surface quality using the Mean Angular Error (MAE) and Cosine Similarity (SIM) of surface normals, which effectively circumvent the errors caused by the spatial misalignment. The detailed computation pipeline is provided in the supplementary material. Furthermore, consistent with Hunyuan3D 2.1, we adopt Uni3D-I~\cite{zhou2023uni3d} and ULIP-I~\cite{xue2023ulip} to assess the semantic and structural alignment between the generated mesh and the GT single-view image.

\subsection{Results and Comparisons}

\subsubsection{Comparison on Pose Canonicalization}
Pose canonicalization—transforming an in-the-wild image with an arbitrary pose into a 3D mesh in a standard canonical pose (e.g., A-pose or T-pose)—is a fundamental sub-task inherently supported by PoseMaster. We compare our approach with task-specific pipelines, CharacterGen~\cite{peng2024charactergen} and StdGen~\cite{he2024stdgen}. Additionally, we include state-of-the-art 3D native generators such as CraftsMan~\cite{li2024craftsman}, Trellis~\cite{xiang2025trellis}, and Hunyuan3D 2.1~\cite{Hunyuan3D2025Hunyuan3D21}. Since these 3D native generators don't support pose stylization, we leverage A-pose images generated from StdGen as their inputs.

To ensure a fair comparison, we evaluate 50 images with diverse poses extracted from the VRoid dataset~\cite{vroid}, which aligns with the training distribution of CharacterGen and StdGen. As reported in Table~\ref{tab:apose}, PoseMaster achieves state-of-the-art performance across all metrics. Qualitatively, Figure~\ref{fig:apose} reveals a critical flaw in cascaded pipelines: the intermediate A-pose images produced by 2D canonicalization methods (e.g., StdGen) frequently suffer from severe structural distortions and artifacts. Due to the strict image-geometry alignment priors of 3D native generators, these 2D artifacts are deterministically projected into the final 3D meshes (such as cases in the first and third rows of Figure~\ref{fig:apose}), leading to significant error accumulation. In contrast, by bypassing the error-prone 2D canonicalization step and directly leveraging 3D skeleton guidance within a native generation workflow, PoseMaster dramatically preserves the identity of the input and yields highly reliable geometry.

\subsubsection{Comparison on Arbitrary-pose Stylization}

\begin{table}[t]
\caption{The quantitative comparison for pose canonicalization on an arbitrary-pose image from the VRoid~\cite{vroid} test dataset. ``*'' denotes that these methods leverage the image-based pose canonicalization method to obtain the A-pose image input.}
\footnotesize
\setlength{\tabcolsep}{5.5pt}
\centering
\begin{tabular}{c|cccc}
\toprule
{Method} & MAE $\downarrow$ & Sim $\uparrow$ & Uni3D-I $\uparrow$ & ULIP-I $\uparrow$\\
\midrule
CharacterGen~\cite{peng2024charactergen} & 6.38 & 0.905 & 0.343 & 0.146 \\
StdGen~\cite{he2024stdgen}  & 4.97 & 0.930 & 0.398 & 0.160 \\
Trellis~\cite{xiang2025trellis}* & 5.39 & 0.926 & 0.398 & 0.157\\ 
CraftsMan~\cite{li2024craftsman}*  & 5.67 & 0.924 & 0.378 & 0.139 \\
Hunyuan3D 2.1~\cite{Hunyuan3D2025Hunyuan3D21}* & 5.89 & 0.920 & 0.398 & 0.150 \\ 
\textbf{PoseMaster (Ours)} & \textbf{4.59} & \textbf{0.938} & \textbf{0.402} & \textbf{0.161} \\
\bottomrule
\end{tabular}
\label{tab:apose}
\end{table}

\begin{figure*}[t]
\includegraphics[width=1\linewidth]{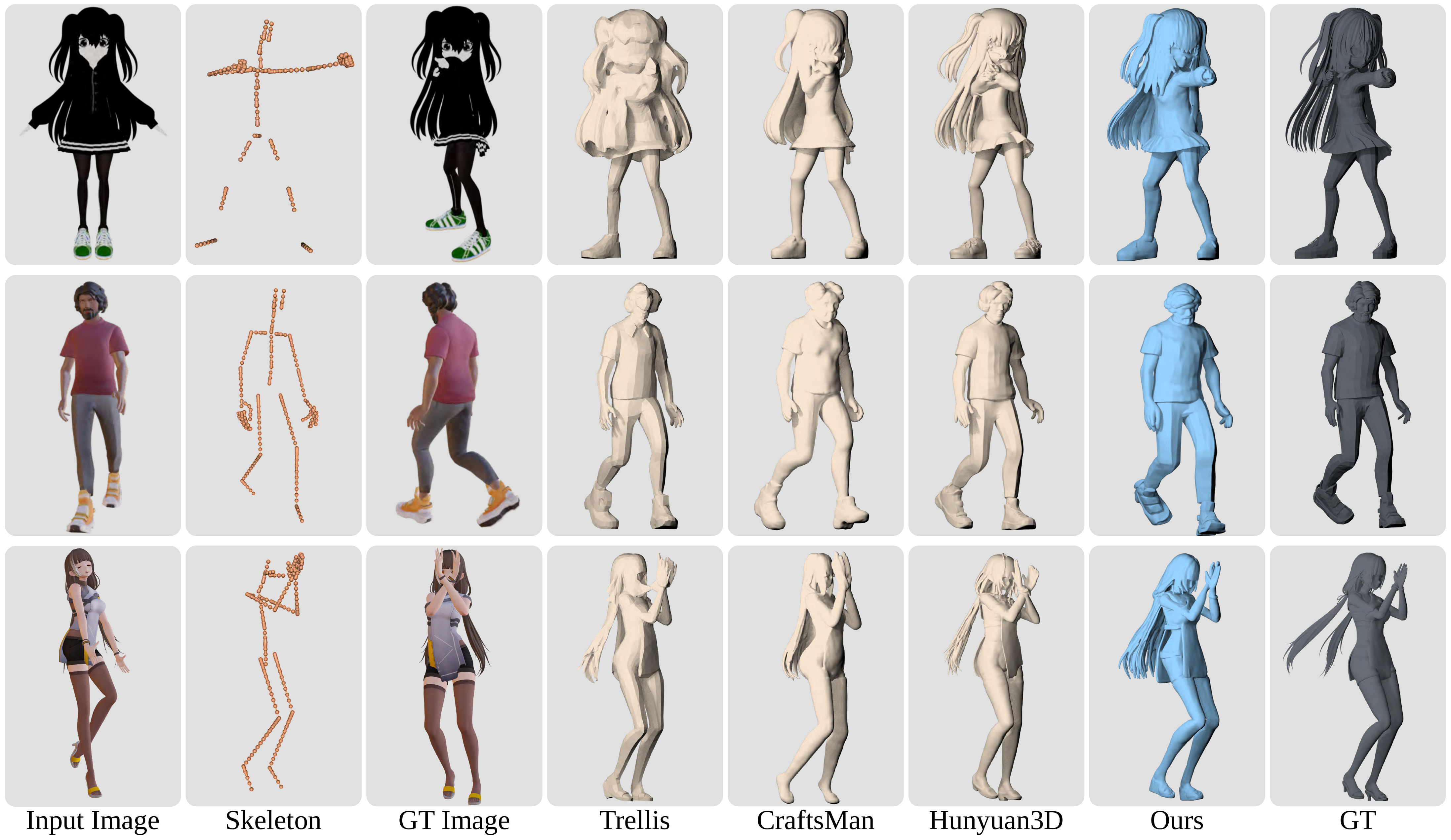}
    \centering
    \caption{Qualitative comparison of arbitrary-pose stylization. For baselines, we utilize the target-pose images (GT image) as the input.}
    \label{fig:arbitrarypose}
\end{figure*}

\begin{table}[t]
\caption{Quantitative comparison for arbitrary-pose stylization from an arbitrary-pose image input. For the compared methods, we directly input the single-view target-pose image while inputting an arbitrary-pose image for our method.}
\footnotesize
\setlength{\tabcolsep}{5.5pt}
\centering
\begin{tabular}{c|ccccc}
\toprule
 {Method} & MAE $\downarrow$ & SIM $\uparrow$ & Uni3D-I $\uparrow$ & ULIP-I $\uparrow$ \\
\midrule
Trellis~\cite{xiang2025trellis} & 7.20 & 0.904 & 0.306 & 0.164\\ 
CraftsMan~\cite{li2024craftsman} & 7.66 & 0.895 & 0.291 & 0.146 \\ 
Hunyuan3D 2.1~\cite{Hunyuan3D2025Hunyuan3D21}  & 6.75 & 0.911 & 0.285 & 0.159 \\
\textbf{PoseMaster (Ours)} & \textbf{5.28} & \textbf{0.935} & \textbf{0.313} & \textbf{0.172} \\
\bottomrule
\end{tabular}
\label{tab:abitrary-pose}
\end{table}

Beyond canonicalization, our method excels at arbitrary pose stylization. However, directly comparing with existing literature is challenging, as 2D image-based methods (e.g., SKDream~\cite{xu2025skdream}, ControlNet~\cite{zhang2023adding}) and video-based methods~\cite{hu2024animate,wang2024unianimate} primarily rely on 2D Openpose-style skeleton as an additional condition for achieving image pose editing, making it difficult to run a fair comparison.

To conduct a rigorous comparison, we establish a highly competitive setting for the 3D native generation methods, including Trellis~\cite{xiang2025trellis}, CraftsMan~\cite{li2024craftsman}, and Hunyuan3D 2.1~\cite{Hunyuan3D2025Hunyuan3D21}. Specifically, we provide these baselines with the target-pose image in our testing dataset, completely eliminating any confounding factors or artifacts arising from 2D pose transfer steps. Meanwhile, our PoseMaster takes the source-pose image and the target-pose 3D skeleton as inputs. Despite the baselines having access to the target-pose image, PoseMaster still demonstrates superior performance in both quantitative and qualitative comparisons, as shown in Table~\ref{tab:abitrary-pose} and Figure~\ref{fig:arbitrarypose}. This outcome underscores the fundamental limitation of purely image-conditioned baselines in resolving topological ambiguities caused by self-occlusion. By explicitly anchoring the generation process to a 3D skeletal prior, PoseMaster reliably deduces coherent global geometry, facilitating topologically robust and complex pose stylizations.



\subsection{Ablation Study} %

\begin{figure}
    \centering
    \includegraphics[width=0.95\linewidth]{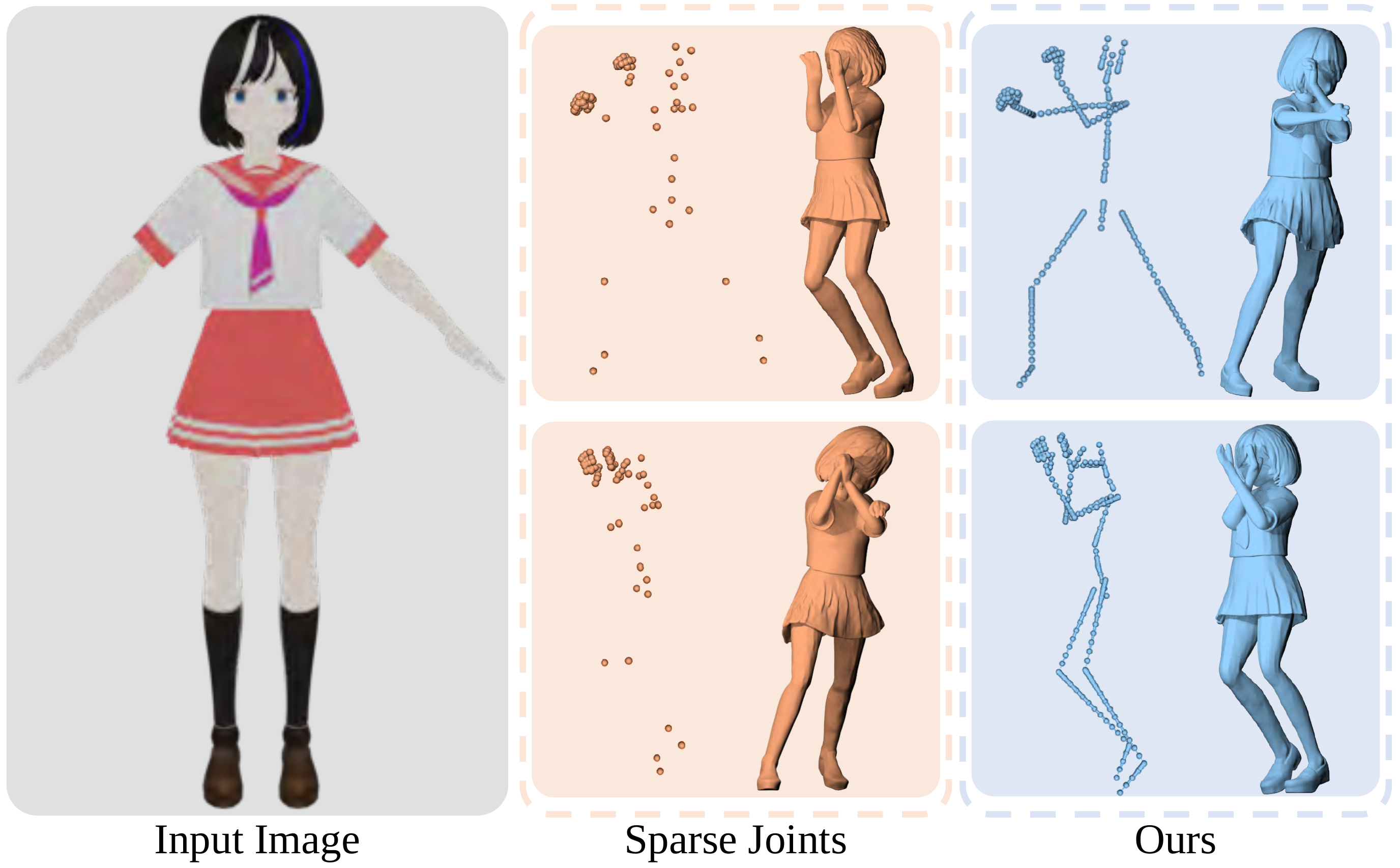}
    \caption{The qualitative comparison between different pose representations in terms of joints and bones of the skeleton.}
    \label{fig:joints}
\end{figure}

\subsubsection{Effect of Pose Representation}

We further investigate the impact of pose representation by comparing conventional sparse joint representations against our proposed dense point clouds. As illustrated in Fig.~\ref{fig:joints}, sparse joints often fail to adequately convey complex articulations, resulting in severe geometric and topological ambiguities. By interpolating the skeletal graph into a dense point cloud augmented with explicit directional vectors, our representation effectively injects robust structural and topological priors into the generation process. This explicitly alleviates the representational burden, enabling the highly accurate reconstruction of complex poses.

\begin{figure}
    \centering
    \includegraphics[width=0.9\linewidth]{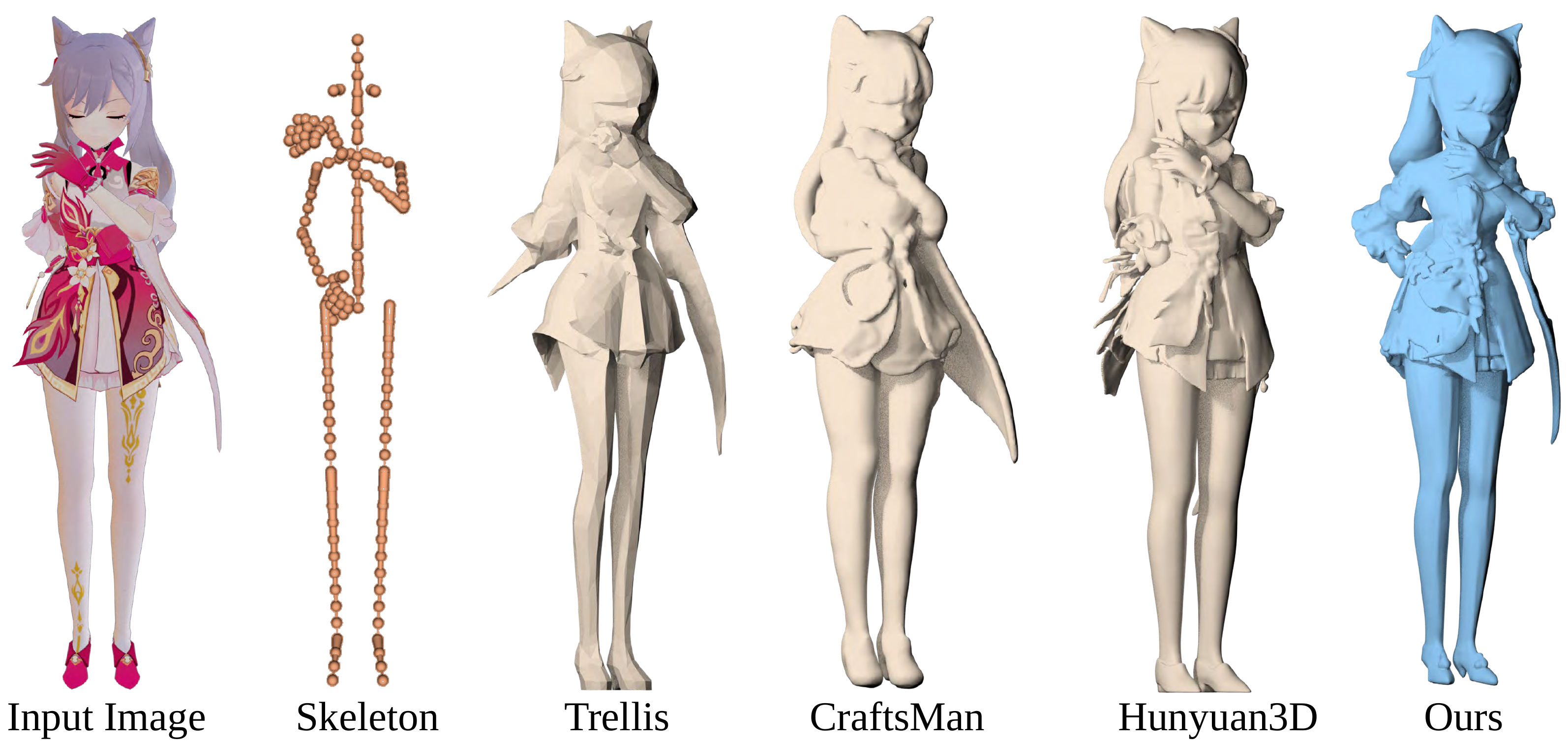}
    \caption{The ablation study for the importance of skeleton guidance in 3D generation.}
    \label{fig:pose-guidance}
\end{figure}

\begin{table}[t]
\caption{Quantitative comparison for skeleton guidance. With the skeleton guidance, our method achieves better performance in the task of image-to-3D generation.}
\footnotesize
\setlength{\tabcolsep}{4.5pt}
\centering
\begin{tabular}{cccccc}
\toprule
 {Method} & MAE $\downarrow$ & SIM $\uparrow$ & Uni3D-I $\uparrow$ & ULIP-I $\uparrow$ \\
\midrule
Trellis~\cite{xiang2025trellis} & 7.53 & 0.898 & 0.303 & 0.153 \\ 
CraftsMan~\cite{li2024craftsman} & 8.01 & 0.893 & 0.304 & 0.152 \\ 
Hunyuan3D 2.1~\cite{Hunyuan3D2025Hunyuan3D21}  & 6.56 & 0.916 & 0.301 & 0.154 \\
\textbf{PoseMaster (Ours)} & \textbf{4.82} & \textbf{0.946} & \textbf{0.315} & \textbf{0.158} \\
\bottomrule
\end{tabular}
\label{tab:pose-guidance}
\end{table}

\subsubsection{Impact of Explicit Skeleton Guidance}
In this ablation study, we isolate and validate the fundamental benefits of explicit 3D skeleton guidance within native 3D generation frameworks. To this end, we provide state-of-the-art native 3D baselines (CraftsMan~\cite{li2024craftsman}, Trellis~\cite{xiang2025trellis}, and Hunyuan3D 2.1~\cite{Hunyuan3D2025Hunyuan3D21}) with only the input image, whereas our PoseMaster is conditioned on the aligned image and 3D skeleton. As reported in Table~\ref{tab:pose-guidance}, our method yields consistently superior geometric accuracy. Furthermore, the qualitative comparisons in Figure~\ref{fig:pose-guidance} confirm that the integrated skeleton acts as a robust geometric anchor in 3D space. This explicit structural prior effectively resolves the depth ambiguity and spatial uncertainty inherent to monocular 3D reconstruction, significantly mitigating geometric collapse, particularly in self-occluded regions.


\section{Application}

\begin{figure}
    \centering
    \includegraphics[width=\linewidth]{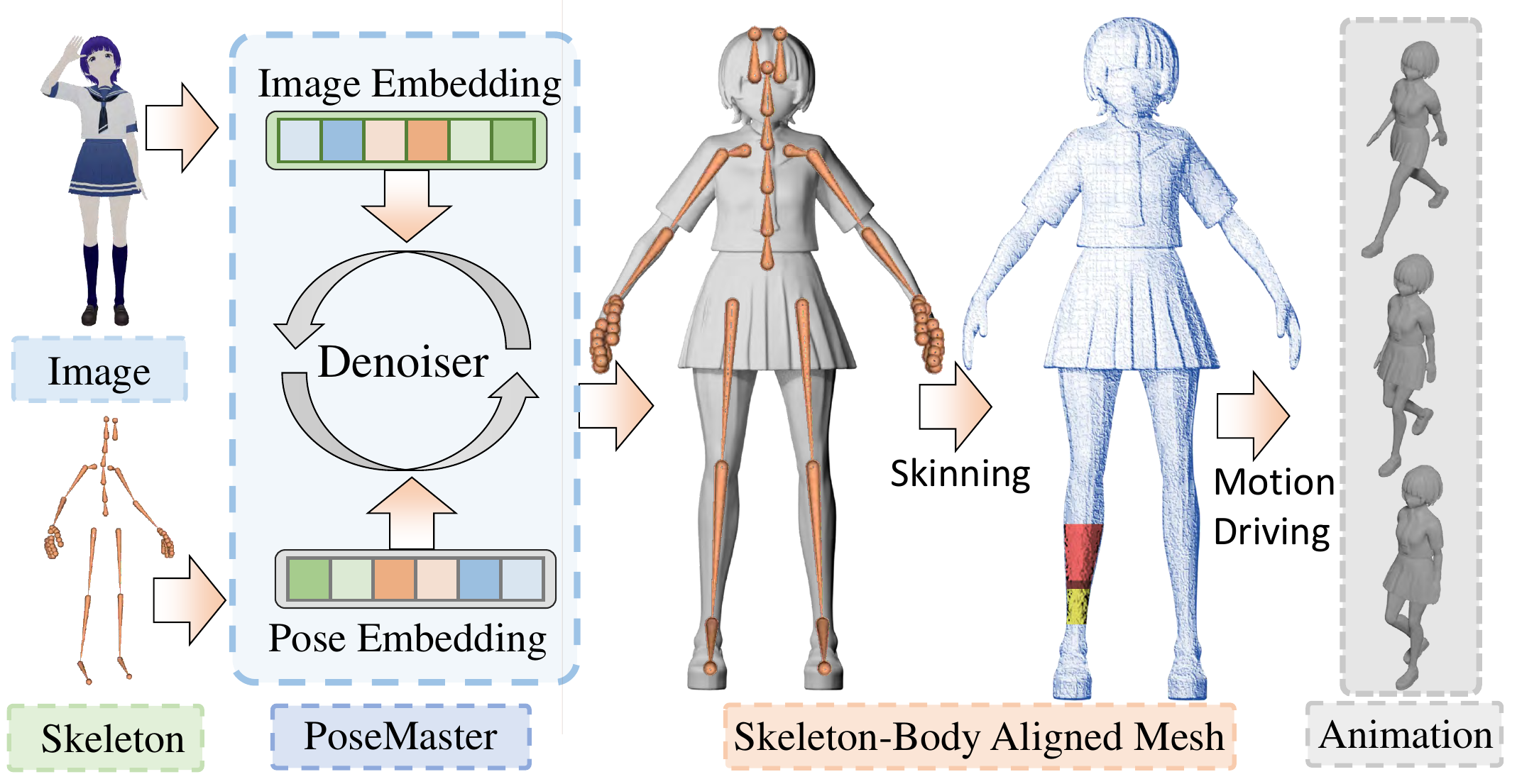}
    \caption{The application of our PoseMaster for animation. Our method can generate the skeleton-body alignment mesh, which can be easily used to animate by introducing a skinning model.}
    \label{fig:animation}
\end{figure}
One compelling application of our method lies in generating rigged meshes to facilitate the creation of animatable 3D assets. Prior rigging approaches, such as UniRig~\cite{unirig} and Puppeteer~\cite{song2025puppeteer}, typically employ auto-regressive (AR) models to predict an underlying skeleton from an input mesh. While these methods successfully generate skeletons that are spatially aligned with the input geometry, they inherently suffer from limited controllability, often struggling to produce the specific skeletal hierarchies desired by users. In contrast, PoseMaster takes a user-specified skeleton as direct input and synthesizes a mesh that strictly conforms to its spatial layout. Consequently, it naturally functions as a highly controllable rigging model, fundamentally circumventing the uncontrollability bottlenecks of the AR paradigm. Furthermore, as illustrated in Figure~\ref{fig:animation}, conditioning the generation on a user-provided skeleton enables seamless integration with standard skinning models, allowing the generated assets to be directly driven by existing motion sequences. This elegantly eliminates the tedious motion retargeting procedures typically required for animating newly generated 3D assets, thereby demonstrating significant practical advantages over prior AR-based methods.

\section{Conclusion}
In this paper, we introduce PoseMaster, a unified native 3D generation framework tailored for arbitrary pose stylization. Unlike conventional cascaded pipelines that rely on 2D foundation models for intermediate pose transfer prior to 3D generation, our approach seamlessly integrates pose stylization and 3D synthesis into a cohesive, end-to-end architecture. This paradigm shift circumvents the limitations of 2D skeleton conditions by directly leveraging explicit 3D skeletons, thereby providing the network with unambiguous spatial constraints and rigorous topological relationships. Consequently, the model's capacity to faithfully synthesize complex target poses is significantly enhanced. Furthermore, we develop a scalable data engine to curate a large-scale, highly diverse dataset composed of pose-decoupled ``Image-Skeleton-Mesh" triplets. Empowered by these innovations, PoseMaster not only establishes a new state-of-the-art in pose canonicalization but also demonstrates unprecedented robustness and versatility in synthesizing high-fidelity 3D assets across arbitrary, in-the-wild poses.
\section{Acknowledgments}
We gratefully acknowledge Xin Huang, Xin Yang, and Bowen Zhang for their assistance with texture generation, and the Tencent IEG VisVise Team for data rendering. We also thank the Tencent Hunyuan 3D Team for their generous support and collaboration throughout this project.

{
    \small
    \bibliographystyle{ieeenat_fullname}
    \bibliography{main}

@String(CVPR= {IEEE Conf. Comput. Vis. Pattern Recog.})

@String(ECCV= {Eur. Conf. Comput. Vis.})

@String(NIPS= {Adv. Neural Inform. Process. Syst.})

@String(TOG= {ACM Trans. Graph.})

@String(ICLR = {Int. Conf. Learn. Represent.})

@String(AAAI = {AAAI})

@String(CVPR  = {CVPR})

@String(ECCV  = {ECCV})

@String(NIPS  = {NeurIPS})

@String(TOG   = {ACM TOG})

@String(ICLR  = {ICLR})

@misc{oquab2023dinov2,
  title={DINOv2: Learning Robust Visual Features without Supervision},
  author={Oquab, Maxime and Darcet, Timothée and Moutakanni, Theo and Vo, Huy V. and Szafraniec, Marc and Khalidov, Vasil and Fernandez, Pierre and Haziza, Daniel and Massa, Francisco and El-Nouby, Alaaeldin and Howes, Russell and Huang, Po-Yao and Xu, Hu and Sharma, Vasu and Li, Shang-Wen and Galuba, Wojciech and Rabbat, Mike and Assran, Mido and Ballas, Nicolas and Synnaeve, Gabriel and Misra, Ishan and Jegou, Herve and Mairal, Julien and Labatut, Patrick and Joulin, Armand and Bojanowski, Piotr},
  journal={arXiv:2304.07193},
  year={2023}
}

@article{wu2024direct3d,
  title={Direct3d: Scalable image-to-3d generation via 3d latent diffusion transformer},
  author={Wu, Shuang and Lin, Youtian and Zhang, Feihu and Zeng, Yifei and Xu, Jingxi and Torr, Philip and Cao, Xun and Yao, Yao},
  journal={arXiv preprint arXiv:2405.14832},
  year={2024}
}

@article{song2025puppeteer,
  title={Puppeteer: Rig and Animate Your 3D Models},
  author={Chaoyue Song and Xiu Li and Fan Yang and Zhongcong Xu and Jiacheng Wei and Fayao Liu and Jiashi Feng and Guosheng Lin and Jianfeng Zhang},
  journal={Advances in Neural Information Processing Systems},
  year={2025}
}

@article{unirig,
author = {Zhang, Jia-Peng and Pu, Cheng-Feng and Guo, Meng-Hao and Cao, Yan-Pei and Hu, Shi-Min},
title = {One Model to Rig Them All: Diverse Skeleton Rigging with UniRig},
year = {2025},
issue_date = {August 2025},
publisher = {Association for Computing Machinery},
address = {New York, NY, USA},
volume = {44},
number = {4},
issn = {0730-0301},
url = {https://doi.org/10.1145/3730930},
doi = {10.1145/3730930},
journal = {ACM Trans. Graph.},
month = jul,
articleno = {123},
numpages = {18}
}

@inproceedings{yan2025symmcompletion,
  title={SymmCompletion: High-fidelity and high-consistency point cloud completion with symmetry guidance},
  author={Yan, Hongyu and Li, Zijun and Luo, Kunming and Lu, Li and Tan, Ping},
  booktitle={Proceedings of the AAAI Conference on Artificial Intelligence},
  volume={39},
  number={9},
  pages={9094--9102},
  year={2025}
}

@inproceedings{yan2022fbnet,
  title={Fbnet: Feedback network for point cloud completion},
  author={Yan, Xuejun and Yan, Hongyu and Wang, Jingjing and Du, Hang and Wu, Zhihong and Xie, Di and Pu, Shiliang and Lu, Li},
  booktitle={European conference on computer vision},
  pages={676--693},
  year={2022},
  organization={Springer}
}

@article{zhang2025easycontrol,
  title={Easycontrol: Adding efficient and flexible control for diffusion transformer},
  author={Zhang, Yuxuan and Yuan, Yirui and Song, Yiren and Wang, Haofan and Liu, Jiaming},
  journal={arXiv preprint arXiv:2503.07027},
  year={2025}
}

@inproceedings{zhang2023adding,
  title={Adding conditional control to text-to-image diffusion models},
  author={Zhang, Lvmin and Rao, Anyi and Agrawala, Maneesh},
  booktitle={Proceedings of the IEEE/CVF international conference on computer vision},
  pages={3836--3847},
  year={2023}
}

@article{zhao2023uni,
  title={Uni-controlnet: All-in-one control to text-to-image diffusion models},
  author={Zhao, Shihao and Chen, Dongdong and Chen, Yen-Chun and Bao, Jianmin and Hao, Shaozhe and Yuan, Lu and Wong, Kwan-Yee K},
  journal={Advances in Neural Information Processing Systems},
  volume={36},
  pages={11127--11150},
  year={2023}
}

@article{peng2024charactergen,
  title={Charactergen: Efficient 3d character generation from single images with multi-view pose canonicalization},
  author={Peng, Hao-Yang and Zhang, Jia-Peng and Guo, Meng-Hao and Cao, Yan-Pei and Hu, Shi-Min},
  journal={ACM Transactions on Graphics (TOG)},
  volume={43},
  number={4},
  pages={1--13},
  year={2024},
  publisher={ACM New York, NY, USA}
}

@inproceedings{22stablediffusion,
  title={High-resolution image synthesis with latent diffusion models},
  author={Rombach, Robin and Blattmann, Andreas and Lorenz, Dominik and Esser, Patrick and Ommer, Bj{\"o}rn},
  booktitle={CVPR},
  year={2022}
}

@article{cao2019openpose,
  title={Openpose: Realtime multi-person 2d pose estimation using part affinity fields},
  author={Cao, Zhe and Hidalgo, Gines and Simon, Tomas and Wei, Shih-En and Sheikh, Yaser},
  journal={IEEE transactions on pattern analysis and machine intelligence},
  volume={43},
  number={1},
  pages={172--186},
  year={2019},
  publisher={IEEE}
}

@article{Zhang_Chen_Yang_Qu_Wang_Chen_Long_Zhu_Du_Zheng_2023,  
 title={AvatarVerse: High-quality \& Stable 3D Avatar Creation from Text and Pose}, 
 author={Zhang, Huichao and Chen, Bowen and Yang, Hao and Qu, Liao and Wang, Xu and Chen, Li and Long, Chao and Zhu, Feida and Du, Kang and Zheng, Min}, 
 year={2023}, 
 month={Aug}, 
 language={en-US} 
 }

@article{cao2023dreamavatar,
  title={Dreamavatar: Text-and-shape guided 3d human avatar generation via diffusion models},
  author={Cao, Yukang and Cao, Yan-Pei and Han, Kai and Shan, Ying and Wong, Kwan-Yee K},
  journal={arXiv preprint arXiv:2304.00916},
  year={2023}
}

@article{long2023wonder3d,
  title={Wonder3d: Single image to 3d using cross-domain diffusion},
  author={Long, Xiaoxiao and Guo, Yuan-Chen and Lin, Cheng and Liu, Yuan and Dou, Zhiyang and Liu, Lingjie and Ma, Yuexin and Zhang, Song-Hai and Habermann, Marc and Theobalt, Christian and others},
  journal={arXiv preprint arXiv:2310.15008},
  year={2023}
}

@article{hong2023lrm,
  title={Lrm: Large reconstruction model for single image to 3d},
  author={Hong, Yicong and Zhang, Kai and Gu, Jiuxiang and Bi, Sai and Zhou, Yang and Liu, Difan and Liu, Feng and Sunkavalli, Kalyan and Bui, Trung and Tan, Hao},
  journal={arXiv preprint arXiv:2311.04400},
  year={2023}
}

@article{liu2022flow,
  title={Flow straight and fast: Learning to generate and transfer data with rectified flow},
  author={Liu, Xingchao and Gong, Chengyue and Liu, Qiang},
  journal={arXiv preprint arXiv:2209.03003},
  year={2022}
}

@article{li2024craftsman,
  title={CraftsMan: High-fidelity Mesh Generation with 3D Native Generation and Interactive Geometry Refiner},
  author={Li, Weiyu and Liu, Jiarui and Chen, Rui and Liang, Yixun and Chen, Xuelin and Tan, Ping and Long, Xiaoxiao},
  journal={arXiv preprint arXiv:2405.14979},
  year={2024}
}

@article{zhang2024clay,
  title={CLAY: A Controllable Large-scale Generative Model for Creating High-quality 3D Assets},
  author={Zhang, Longwen and Wang, Ziyu and Zhang, Qixuan and Qiu, Qiwei and Pang, Anqi and Jiang, Haoran and Yang, Wei and Xu, Lan and Yu, Jingyi},
  journal={ACM Transactions on Graphics (TOG)},
  volume={43},
  number={4},
  pages={1--20},
  year={2024},
  publisher={ACM New York, NY, USA}
}

@article{zhang20233dshape2vecset,
  title={3dshape2vecset: A 3d shape representation for neural fields and generative diffusion models},
  author={Zhang, Biao and Tang, Jiapeng and Niessner, Matthias and Wonka, Peter},
  journal={ACM Transactions on Graphics (TOG)},
  volume={42},
  number={4},
  pages={1--16},
  year={2023},
  publisher={ACM New York, NY, USA}
}

@inproceedings{jiang2023avatarcraft,
  title={Avatarcraft: Transforming text into neural human avatars with parameterized shape and pose control},
  author={Jiang, Ruixiang and Wang, Can and Zhang, Jingbo and Chai, Menglei and He, Mingming and Chen, Dongdong and Liao, Jing},
  booktitle={Proceedings of the IEEE/CVF International Conference on Computer Vision},
  pages={14371--14382},
  year={2023}
}

@article{kolotouros2023dreamhuman,
  title={Dreamhuman: Animatable 3d avatars from text},
  author={Kolotouros, Nikos and Alldieck, Thiemo and Zanfir, Andrei and Bazavan, Eduard and Fieraru, Mihai and Sminchisescu, Cristian},
  journal={Advances in Neural Information Processing Systems},
  volume={36},
  pages={10516--10529},
  year={2023}
}

@article{zhang2025avatarstudio,
  title={Avatarstudio: High-fidelity and animatable 3d avatar creation from text},
  author={Zhang, Xuanmeng and Zhang, Jianfeng and Zhang, Chenxu and Liew, Jun Hao and Zhang, Huichao and Yang, Yi and Feng, Jiashi},
  journal={International Journal of Computer Vision},
  pages={1--19},
  year={2025},
  publisher={Springer}
}

@inproceedings{liao2024tada,
  title={Tada! text to animatable digital avatars},
  author={Liao, Tingting and Yi, Hongwei and Xiu, Yuliang and Tang, Jiaxiang and Huang, Yangyi and Thies, Justus and Black, Michael J},
  booktitle={2024 International Conference on 3D Vision (3DV)},
  pages={1508--1519},
  year={2024},
  organization={IEEE}
}

@article{he2024stdgen,
  title={StdGEN: Semantic-Decomposed 3D Character Generation from Single Images},
  author={He, Yuze and Zhou, Yanning and Zhao, Wang and Wu, Zhongkai and Xiao, Kaiwen and Yang, Wei and Liu, Yong-Jin and Han, Xiao},
  journal={arXiv preprint arXiv:2411.05738},
  year={2024}
}

@article{li2024era3d,
  title={Era3D: High-Resolution Multiview Diffusion using Efficient Row-wise Attention},
  author={Li, Peng and Liu, Yuan and Long, Xiaoxiao and Zhang, Feihu and Lin, Cheng and Li, Mengfei and Qi, Xingqun and Zhang, Shanghang and Luo, Wenhan and Tan, Ping and others},
  journal={arXiv preprint arXiv:2405.11616},
  year={2024}
}

@inproceedings{ren2024xcube,
    title={XCube: Large-Scale 3D Generative Modeling using Sparse Voxel Hierarchies}, 
    author={Ren, Xuanchi and Huang, Jiahui and Zeng, Xiaohui and Museth, Ken and Fidler, Sanja and Williams, Francis},
    booktitle={Proceedings of the IEEE/CVF Conference on Computer Vision and Pattern Recognition},
    year={2024}
}

@inproceedings{zhao2023michelangelo,
title={Michelangelo: Conditional 3D Shape Generation based on Shape-Image-Text Aligned Latent Representation},
author={Zibo Zhao and Wen Liu and Xin Chen and Xianfang Zeng and Rui Wang and Pei Cheng and BIN FU and Tao Chen and Gang YU and Shenghua Gao},
booktitle=NIPS,
year={2023},
url={https://openreview.net/forum?id=xmxgMij3LY}
}

@inproceedings{chen2019learning,
  title={Learning implicit fields for generative shape modeling},
  author={Chen, Zhiqin and Zhang, Hao},
  booktitle=CVPR,
  pages={5939--5948},
  year={2019}
}

@inproceedings{park2019deepsdf,
  title={Deepsdf: Learning continuous signed distance functions for shape representation},
  author={Park, Jeong Joon and Florence, Peter and Straub, Julian and Newcombe, Richard and Lovegrove, Steven},
  booktitle=CVPR,
  pages={165--174},
  year={2019}
}

@article{li2018point,
  title={Point cloud gan},
  author={Li, Chun-Liang and Zaheer, Manzil and Zhang, Yang and Poczos, Barnabas and Salakhutdinov, Ruslan},
  journal={arXiv preprint arXiv:1810.05795},
  year={2018}
}

@article{pointflow,
 title={PointFlow: 3D Point Cloud Generation with Continuous Normalizing Flows},
 author={Yang, Guandao and Huang, Xun and Hao, Zekun and Liu, Ming-Yu and Belongie, Serge and Hariharan, Bharath},
 journal={arXiv},
 year={2019}
}

@inproceedings{zhou20213d,
  title={3d shape generation and completion through point-voxel diffusion},
  author={Zhou, Linqi and Du, Yilun and Wu, Jiajun},
  booktitle={Proceedings of the IEEE/CVF international conference on computer vision},
  pages={5826--5835},
  year={2021}
}

@inproceedings{nash2020polygen,
  title={Polygen: An autoregressive generative model of 3d meshes},
  author={Nash, Charlie and Ganin, Yaroslav and Eslami, SM Ali and Battaglia, Peter},
  booktitle={International conference on machine learning},
  pages={7220--7229},
  year={2020},
  organization={PMLR}
}

@inproceedings{Liu2023MeshDiffusion,
    title={MeshDiffusion: Score-based Generative 3D Mesh Modeling},
    author={Zhen Liu and Yao Feng and Michael J. Black and Derek Nowrouzezahrai and Liam Paull and Weiyang Liu},
    booktitle={International Conference on Learning Representations},
    year={2023},
    url={https://openreview.net/forum?id=0cpM2ApF9p6}
}

@inproceedings{siddiqui2024meshgpt,
  title={Meshgpt: Generating triangle meshes with decoder-only transformers},
  author={Siddiqui, Yawar and Alliegro, Antonio and Artemov, Alexey and Tommasi, Tatiana and Sirigatti, Daniele and Rosov, Vladislav and Dai, Angela and Nie{\ss}ner, Matthias},
  booktitle={Proceedings of the IEEE/CVF conference on computer vision and pattern recognition},
  pages={19615--19625},
  year={2024}
}

@Misc{playbox,
title = {Playbox},
note = {https://www.aplaybox.com},
year = 2025,
}

@inproceedings{liuLQWTcvpr16DeepFashion,
 author = {Liu, Ziwei and Luo, Ping and Qiu, Shi and Wang, Xiaogang and Tang, Xiaoou},
 title = {DeepFashion: Powering Robust Clothes Recognition and Retrieval with Rich Annotations},
 booktitle = {Proceedings of IEEE Conference on Computer Vision and Pattern Recognition (CVPR)},
 month = {June},
 year = {2016}
 }

@article{qi2017pointnet++,
  title={Pointnet++: Deep hierarchical feature learning on point sets in a metric space},
  author={Qi, Charles Ruizhongtai and Yi, Li and Su, Hao and Guibas, Leonidas J},
  journal={Advances in neural information processing systems},
  volume={30},
  year={2017}
}

@inproceedings{xue2023ulip,
  title={Ulip: Learning a unified representation of language, images, and point clouds for 3d understanding},
  author={Xue, Le and Gao, Mingfei and Xing, Chen and Mart{\'\i}n-Mart{\'\i}n, Roberto and Wu, Jiajun and Xiong, Caiming and Xu, Ran and Niebles, Juan Carlos and Savarese, Silvio},
  booktitle={Proceedings of the IEEE/CVF conference on computer vision and pattern recognition},
  pages={1179--1189},
  year={2023}
}

@article{li2025sparc3d,
  title={Sparc3D: Sparse Representation and Construction for High-Resolution 3D Shapes Modeling},
  author={Li, Zhihao and Wang, Yufei and Zheng, Heliang and Luo, Yihao and Wen, Bihan},
  journal={arXiv preprint arXiv:2505.14521},
  year={2025}
}

@article{zhou2023uni3d,
  title={Uni3d: Exploring unified 3d representation at scale},
  author={Zhou, Junsheng and Wang, Jinsheng and Ma, Baorui and Liu, Yu-Shen and Huang, Tiejun and Wang, Xinlong},
  journal={arXiv preprint arXiv:2310.06773},
  year={2023}
}

@article{ke2025marigold,
  title={Marigold: Affordable Adaptation of Diffusion-Based Image Generators for Image Analysis},
  author={Ke, Bingxin and Qu, Kevin and Wang, Tianfu and Metzger, Nando and Huang, Shengyu and Li, Bo and Obukhov, Anton and Schindler, Konrad},
  journal={arXiv preprint arXiv:2505.09358},
  year={2025}
}

@article{lai2025hunyuan3d,
  title={Hunyuan3d 2.5: Towards high-fidelity 3d assets generation with ultimate details},
  author={Lai, Zeqiang and Zhao, Yunfei and Liu, Haolin and Zhao, Zibo and Lin, Qingxiang and Shi, Huiwen and Yang, Xianghui and Yang, Mingxin and Yang, Shuhui and Feng, Yifei and others},
  journal={arXiv preprint arXiv:2506.16504},
  year={2025}
}

@article{lai2025natex,
  title={NaTex: Seamless Texture Generation as Latent Color Diffusion},
  author={Lai, Zeqiang and Zhao, Yunfei and Zhao, Zibo and Yang, Xin and Huang, Xin and Huang, Jingwei and Yue, Xiangyu and Guo, Chunchao},
  journal={arXiv preprint arXiv:2511.16317},
  year={2025}
}

@article{chen2025lafite,
  title={LaFiTe: A Generative Latent Field for 3D Native Texturing},
  author={Chen, Chia-Hao and Zou, Zi-Xin and Cao, Yan-Pei and Yuan, Ze and Luo, Guan and Qi, Xiaojuan and Liang, Ding and Zhang, Song-Hai and Guo, Yuan-Chen},
  journal={arXiv preprint arXiv:2512.04786},
  year={2025}
}

@inproceedings{fu2024geowizard,
  title={Geowizard: Unleashing the diffusion priors for 3d geometry estimation from a single image},
  author={Fu, Xiao and Yin, Wei and Hu, Mu and Wang, Kaixuan and Ma, Yuexin and Tan, Ping and Shen, Shaojie and Lin, Dahua and Long, Xiaoxiao},
  booktitle={European Conference on Computer Vision},
  pages={241--258},
  year={2024},
  organization={Springer}
}

@inproceedings{xu2025skdream,
  title={SKDream: Controllable Multi-view and 3D Generation with Arbitrary Skeletons},
  author={Xu, Yuanyou and Yang, Zongxin and Yang, Yi},
  booktitle={Proceedings of the Computer Vision and Pattern Recognition Conference},
  pages={314--325},
  year={2025}
}

@Misc{vroid,
title = {VRoid-Hub},
note = {https://hub.vroid.com/},
year = 2025,
}

@Misc{GPT4o,
title = {GPT4-o},
note = {https://openai.com/index/hello-gpt-4o/},
year = 2025,
}

@Misc{Nano,
title = {NanoBanana},
note = {https://aistudio.google.com/models/gemini-2-5-flash-image},
year = 2025,
}

@Misc{seed,
title = {SEED},
note = {https://www.doubao.com/},
year = 2025,
}

@Misc{Readyplayerme,
title = {Readyplayerme},
note = {https://readyplayer.me/},
year = 2025,
}

@inproceedings{deitke2023objaverse,
  title={Objaverse: A universe of annotated 3d objects},
  author={Deitke, Matt and Schwenk, Dustin and Salvador, Jordi and Weihs, Luca and Michel, Oscar and VanderBilt, Eli and Schmidt, Ludwig and Ehsani, Kiana and Kembhavi, Aniruddha and Farhadi, Ali},
  booktitle={Proceedings of the IEEE/CVF conference on computer vision and pattern recognition},
  pages={13142--13153},
  year={2023}
}

@article{deitke2023objaversexl,
  title={Objaverse-xl: A universe of 10m+ 3d objects},
  author={Deitke, Matt and Liu, Ruoshi and Wallingford, Matthew and Ngo, Huong and Michel, Oscar and Kusupati, Aditya and Fan, Alan and Laforte, Christian and Voleti, Vikram and Gadre, Samir Yitzhak and others},
  journal={Advances in Neural Information Processing Systems},
  volume={36},
  pages={35799--35813},
  year={2023}
}

@inproceedings{chu2025humanrig,
  title={Humanrig: Learning automatic rigging for humanoid character in a large scale dataset},
  author={Chu, Zedong and Xiong, Feng and Liu, Meiduo and Zhang, Jinzhi and Shao, Mingqi and Sun, Zhaoxu and Wang, Di and Xu, Mu},
  booktitle={Proceedings of the Computer Vision and Pattern Recognition Conference},
  pages={304--313},
  year={2025}
}

@inproceedings{hu2024animate,
  title={Animate anyone: Consistent and controllable image-to-video synthesis for character animation},
  author={Hu, Li},
  booktitle={Proceedings of the IEEE/CVF Conference on Computer Vision and Pattern Recognition},
  pages={8153--8163},
  year={2024}
}

@inproceedings{zhang2024joint2human,
  title={Joint2human: High-quality 3d human generation via compact spherical embedding of 3d joints},
  author={Zhang, Muxin and Feng, Qiao and Su, Zhuo and Wen, Chao and Xue, Zhou and Li, Kun},
  booktitle={Proceedings of the IEEE/CVF Conference on Computer Vision and Pattern Recognition},
  pages={1429--1438},
  year={2024}
}

@misc{xu2025hypermotion,
    title={HyperMotion: DiT-Based Pose-Guided Human Image Animation of Complex Motions}, 
    author={Shuolin Xu and Siming Zheng and Ziyi Wang and HC Yu and Jinwei Chen and Huaqi Zhang and Bo Li and Peng-Tao Jiang},
    year={2025},
    eprint={2505.22977},
    archivePrefix={arXiv},
    primaryClass={cs.CV},
    url={https://arxiv.org/abs/2505.22977}, 
  }

@article{tan2024animate,
  title={Animate-x: Universal character image animation with enhanced motion representation},
  author={Tan, Shuai and Gong, Biao and Wang, Xiang and Zhang, Shiwei and Zheng, Dandan and Zheng, Ruobing and Zheng, Kecheng and Chen, Jingdong and Yang, Ming},
  journal={arXiv preprint arXiv:2410.10306},
  year={2024}
}

@article{wang2024unianimate,
      title={UniAnimate: Taming Unified Video Diffusion Models for Consistent Human Image Animation},
      author={Wang, Xiang and Zhang, Shiwei and Gao, Changxin and Wang, Jiayu and Zhou, Xiaoqiang and Zhang, Yingya and Yan, Luxin and Sang, Nong},
      journal={arXiv preprint arXiv:2406.01188},
      year={2024}
}

@article{wu2025qwen,
  title={Qwen-image technical report},
  author={Wu, Chenfei and Li, Jiahao and Zhou, Jingren and Lin, Junyang and Gao, Kaiyuan and Yan, Kun and Yin, Sheng-ming and Bai, Shuai and Xu, Xiao and Chen, Yilei and others},
  journal={arXiv preprint arXiv:2508.02324},
  year={2025}
}

@article{xie2023omnicontrol,
  title={Omnicontrol: Control any joint at any time for human motion generation},
  author={Xie, Yiming and Jampani, Varun and Zhong, Lei and Sun, Deqing and Jiang, Huaizu},
  journal={arXiv preprint arXiv:2310.08580},
  year={2023}
}

@article{peng2024controlnext,
  title={Controlnext: Powerful and efficient control for image and video generation},
  author={Peng, Bohao and Wang, Jian and Zhang, Yuechen and Li, Wenbo and Yang, Ming-Chang and Jia, Jiaya},
  journal={arXiv preprint arXiv:2408.06070},
  year={2024}
}

@Misc{blender,
title = {Blender},
note = {https://www.blender.org/},
year = 2025,
}

@article{hunyuan3d2025hunyuan3d21,
  title={Hunyuan3D 2.1: From Images to High-Fidelity 3D Assets with Production-Ready PBR Material},
  author={Hunyuan3D, Team and Yang, Shuhui and Yang, Mingxin and Feng, Yifei and Huang, Xin and Zhang, Sheng and He, Zebin and Luo, Di and Liu, Haolin and Zhao, Yunfei and others},
  journal={arXiv preprint arXiv:2506.15442},
  year={2025}
}

@inproceedings{xiang2025trellis,
  title={Structured 3d latents for scalable and versatile 3d generation},
  author={Xiang, Jianfeng and Lv, Zelong and Xu, Sicheng and Deng, Yu and Wang, Ruicheng and Zhang, Bowen and Chen, Dong and Tong, Xin and Yang, Jiaolong},
  booktitle={Proceedings of the Computer Vision and Pattern Recognition Conference},
  pages={21469--21480},
  year={2025}
}

@article{chen2025ultra3d,
  title={Ultra3d: Efficient and high-fidelity 3d generation with part attention},
  author={Chen, Yiwen and Li, Zhihao and Wang, Yikai and Zhang, Hu and Li, Qin and Zhang, Chi and Lin, Guosheng},
  journal={arXiv preprint arXiv:2507.17745},
  year={2025}
}

@article{direct3ds2,
  author       = {Shuang Wu and
                  Youtian Lin and
                  Feihu Zhang and
                  Yifei Zeng and
                  Yikang Yang and
                  Yajie Bao and
                  Jiachen Qian and
                  Siyu Zhu and
                  Xun Cao and
                  Philip Torr and
                  Yao Yao},
  title        = {Direct3D-S2: Gigascale 3D Generation Made Easy with Spatial Sparse
                  Attention},
  journal      = {CoRR},
  volume       = {abs/2505.17412},
  year         = {2025},
  url          = {https://doi.org/10.48550/arXiv.2505.17412},
  doi          = {10.48550/ARXIV.2505.17412},
  eprinttype    = {arXiv},
  eprint       = {2505.17412},
  bibsource    = {dblp computer science bibliography, https://dblp.org}
}

@article{triposg,
  author       = {Yangguang Li and
                  Zi{-}Xin Zou and
                  Zexiang Liu and
                  Dehu Wang and
                  Yuan Liang and
                  Zhipeng Yu and
                  Xingchao Liu and
                  Yuan{-}Chen Guo and
                  Ding Liang and
                  Wanli Ouyang and
                  Yan{-}Pei Cao},
  title        = {TripoSG: High-Fidelity 3D Shape Synthesis using Large-Scale Rectified
                  Flow Models},
  journal      = {CoRR},
  volume       = {abs/2502.06608},
  year         = {2025},
  url          = {https://doi.org/10.48550/arXiv.2502.06608},
  doi          = {10.48550/ARXIV.2502.06608},
  eprinttype    = {arXiv},
  eprint       = {2502.06608},
  bibsource    = {dblp computer science bibliography, https://dblp.org}
}

@inproceedings{lan2024ln3diff,
    title={LN3Diff: Scalable Latent Neural Fields Diffusion for Speedy 3D Generation}, 
    author={Lan, Yushi and Hong, Fangzhou and Yang, Shuai and Zhou, Shangchen and Meng, Xuyi and Dai, Bo and Pan, Xingang and Loy, Chen Change},
    year={2024},
    booktitle={ECCV},
}

@inproceedings{GaussianAnything,
  title={GaussianAnything: Interactive Point Cloud Latent Diffusion for 3D Generation},
  author={Lan, Yushi and Zhou, Shangchen and Lyu, Zhaoyang and Hong, Fangzhou and Yang, Shuai and Dai, Bo and Pan, Xingang and Loy, Chen Change},
  year={2025},
  booktitle={ICLR}
}

@article{3DTopia-XL,
    title={3DTopia-XL: High-Quality 3D PBR Asset Generation via Primitive Diffusion},
    author={Chen, Zhaoxi and Tang, Jiaxiang and Dong, Yuhao and Cao, Ziang and Hong, Fangzhou and Lan, Yushi and Wang, Tengfei and Xie, Haozhe and Wu, Tong and Saito, Shunsuke and Pan, Liang and Lin, Dahua and Liu, Ziwei},
    journal={arXiv preprint arXiv:2409.12957},
    year={2024}
}

@article{stepfun3d,
  title={Step1X-3D: Towards High-Fidelity and Controllable Generation of Textured 3D Assets},
  author={Li, Weiyu and Zhang, Xuanyang and Sun, Zheng and Qi, Di and Li, Hao and Cheng, Wei and Cai, Weiwei and Wu, Shihao and Liu, Jiarui and Wang, Zihao and others},
  journal={arXiv preprint arXiv:2505.07747},
  year={2025}
}

@inproceedings{
anonymous2025voxset,
title={VoxSet: Sparse Voxel Set Tokenizer for 3D Shape Generation},
author={Anonymous},
booktitle={Submitted to The Fourteenth International Conference on Learning Representations},
year={2025},
url={https://openreview.net/forum?id=7cLvFw1ZGu},
note={under review}
}

@article{ye2025hi3dgen,
  title={Hi3DGen: High-fidelity 3D Geometry Generation from Images via Normal Bridging},
  author={Ye, Chongjie and Wu, Yushuang and Lu, Ziteng and Chang, Jiahao and Guo, Xiaoyang and Zhou, Jiaqing and Zhao, Hao and Han, Xiaoguang},
  journal={arXiv preprint arXiv:2503.22236}, 
  year={2025}
}

@article{liang2025unitex,
  title={Unitex: Universal high fidelity generative texturing for 3d shapes},
  author={Liang, Yixun and Luo, Kunming and Chen, Xiao and Chen, Rui and Yan, Hongyu and Li, Weiyu and Liu, Jiarui and Tan, Ping},
  journal={arXiv preprint arXiv:2505.23253},
  year={2025}
}

@inproceedings{luoctr3d,
  title={CTR3D: Cross-view Token Reduction for Dense Multi-view Generation},
  author={Luo, Kunming and Yan, Hongyu and Liu, Yuan and Zhang, Zihao and Zhang, Manyuan and Wang, Wenping and Tan, Ping},
  booktitle={Thirteenth International Conference on 3D Vision}
}
}


\section{Additional Details}
In this section, we provide additional details to clarify our method, including metric computation, data details, and skeleton definition.

\subsection{Metric Computation}
Given the sensitivity of Chamfer Distance (CD) to shape alignment and point distribution—particularly the topological discrepancies between watertight and non-watertight meshes—we adopt Mean Angular Error (MAE) and Cosine Similarity (SIM) of rendered normal maps to assess geometric accuracy. Specifically, our evaluation begins by rendering the front-view normal map of the predicted mesh. To account for potential azimuthal misalignments between the predicted meshes (including those generated by baseline methods such as Hunyuan3D 2.1~\cite{Hunyuan3D2025Hunyuan3D21} and CraftsMan~\cite{li2024craftsman}) and the ground truth (GT) models, we render the GT normals across 36 azimuths uniformly sampled at $10^\circ$ intervals (from $0^\circ$ to $350^\circ$). We then compute the MAE and SIM between the predicted front-view normals and the GT normals under each viewpoint. The best-matching score among all views is reported as the final metric, ensuring a robust evaluation of geometric fidelity.

\subsection{Data details.}
Our constructed dataset comprises animatable dynamic assets and static assets in an approximate 3:2 ratio. For static meshes lacking skeletal structures, we employ an auto-regressive model fine-tuned from UniRig~\cite{unirig} for automatic rig generation. For the dynamic assets, the characters from ReadyPlayerMe~\cite{Readyplayerme} and VRoid~\cite{vroid} are driven by Mixamo motion sequences, whereas the assets from Playbox~\cite{playbox} are animated by VMD motion sequences compatible with the MMD data format.

{\bf Image rendering.} For dynamic characters, we fix the camera azimuth, allowing the character's articulated motions to naturally induce viewpoint variations. Furthermore, we randomize the camera elevation between $0^\circ$ and $10^\circ$ and dynamically adjust the camera distance to ensure the animated character remains centered in the viewpoint. For static meshes, we render four canonical views (front, back, left, and right) alongside 32 uniformly sampled random views. All renderings are generated using perspective projection.

\subsection{Skeleton Definition}
\begin{figure}[t]
\includegraphics[width=1\linewidth]{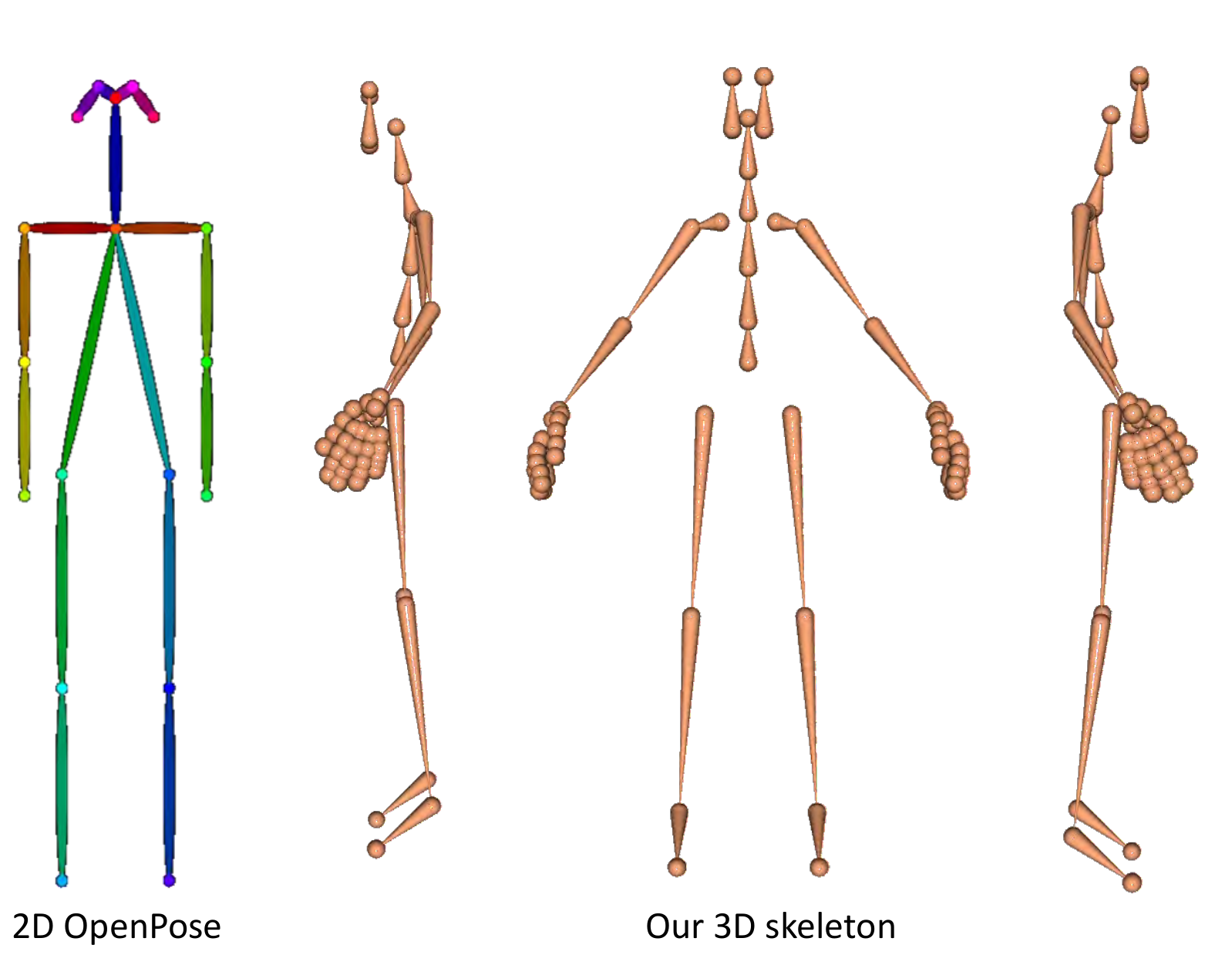}
    \centering
    \caption{The visualization of the standard skeletons of Openpose~\cite{cao2019Openpose} and ours (here, we take the skeleton from Readyplayeme's~\cite{Readyplayerme} data as the example).}
    \label{fig:charactergen}
\end{figure}
Following previous methods~\cite{peng2024charactergen,tan2024animate} that utilize Openpose~\cite{cao2019Openpose} for pose conditioning, we select the body's bones and hand's bones in the 3D skeleton as our skeleton system. This selection excludes hair and skirt bones, yet remains sufficient to define a humanoid pose effectively.

\section{Additional Experimental Results}
In this section, we present extended comparative evaluations, featuring additional qualitative visualizations for both arbitrary-pose stylization and pose canonicalization. Following these comparisons, we provide a comprehensive analysis of the computational efficiency and structural robustness of our proposed framework. Finally, we report supplementary texture synthesis results and demonstrate a practical downstream application.

\subsection{More results on Arbitrary-pose Stylization}
\begin{figure*}[t]
\includegraphics[width=0.95\linewidth]{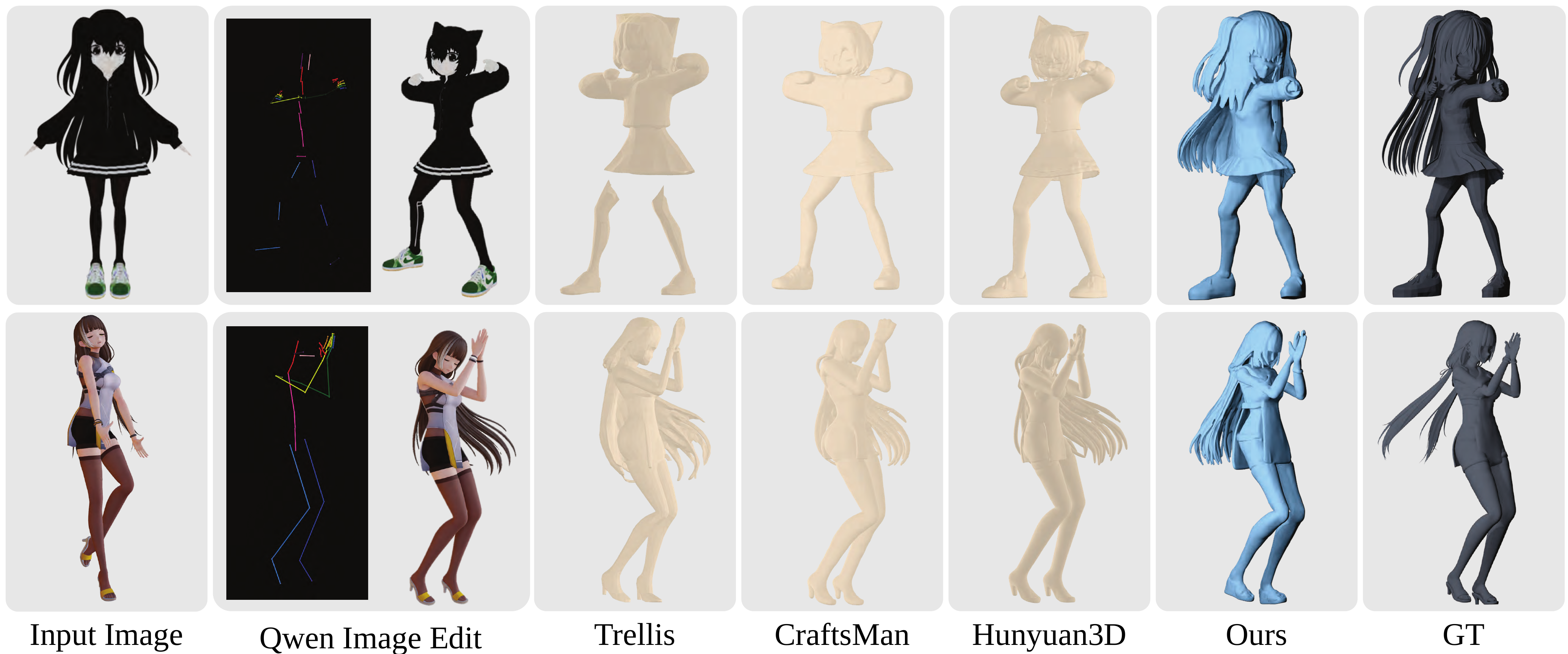}
    \centering
    \caption{The qualitative comparison for arbitrary-pose stylization by using the target-pose images edited from Qwen-Image~\cite{wu2025qwen} as the baselines' input.}
    \label{fig:full_arbitrary}
\end{figure*}

\begin{table}[t]
\caption{The quantitative comparison for arbitrary pose stylization by using the target-pose images edited from Qwen-Image~\cite{wu2025qwen} as the baselines' input.}
\footnotesize
\setlength{\tabcolsep}{5.5pt}
\centering
\begin{tabular}{c|ccccc}
\toprule
 {Method} & MAE $\downarrow$ & SIM $\uparrow$ & Uni3D-I $\uparrow$ & ULIP-I $\uparrow$ \\
\midrule
Trellis~\cite{xiang2025trellis} & 8.50 & 0.874 & 0.293 & 0.160\\ 
CraftsMan~\cite{li2024craftsman} & 8.55 & 0.876 & 0.302 & 0.148 \\ 
Hunyuan3D 2.1~\cite{Hunyuan3D2025Hunyuan3D21}  & 8.15 & 0.874 & 0.293 & 0.160 \\
\textbf{PoseMaster (Ours)} & \textbf{5.28} & \textbf{0.935} & \textbf{0.313} & \textbf{0.172} \\
\bottomrule
\end{tabular}
\label{tab:full_arbitrary}
\end{table}

To isolate the effects of 2D image distortion inherent in decoupled pose stylization pipelines, we utilized ground-truth target-pose images as inputs for existing 3D native methods (i.e., Trellis~\cite{xiang2025trellis}, CraftsMan~\cite{li2024craftsman}, and Hunyuan3D 2.1~\cite{Hunyuan3D2025Hunyuan3D21}) in the arbitrary-pose stylization comparisons presented in Sec.~5.2.2. Therefore, to explicitly expose the limitations of the decoupled paradigm—where 2D pose transformation and 3D generation are treated as independent stages—we construct a comprehensive baseline pipeline utilizing Qwen-Image~\cite{wu2025qwen} to synthesize target-pose inputs. Specifically, we project 3D skeletons into Openpose-style 2D skeleton maps to guide Qwen-Image in generating pose-edited images, which are subsequently fed into 3D generation models to produce meshes.

As shown in Table~\ref{tab:full_arbitrary} and Figure~\ref{fig:full_arbitrary}, our method presents significant quantitative and qualitative superiority over both baseline methods. Visual inspection reveals that Qwen-Image-based editing frequently compromises the preservation of original identity features. Moreover, the 2D interpretation of skeletons suffers from viewpoint-dependent ambiguity, leading to geometric inaccuracies in the generated poses. Consequently, the performance of the Qwen-Image-based pipeline falls short of the setting that uses ground-truth inputs. This performance gap strongly underscores the critical necessity of a unified framework and explicit 3D skeleton guidance for robust 3D pose stylization.

\begin{figure*}[t]
\includegraphics[width=1\linewidth]{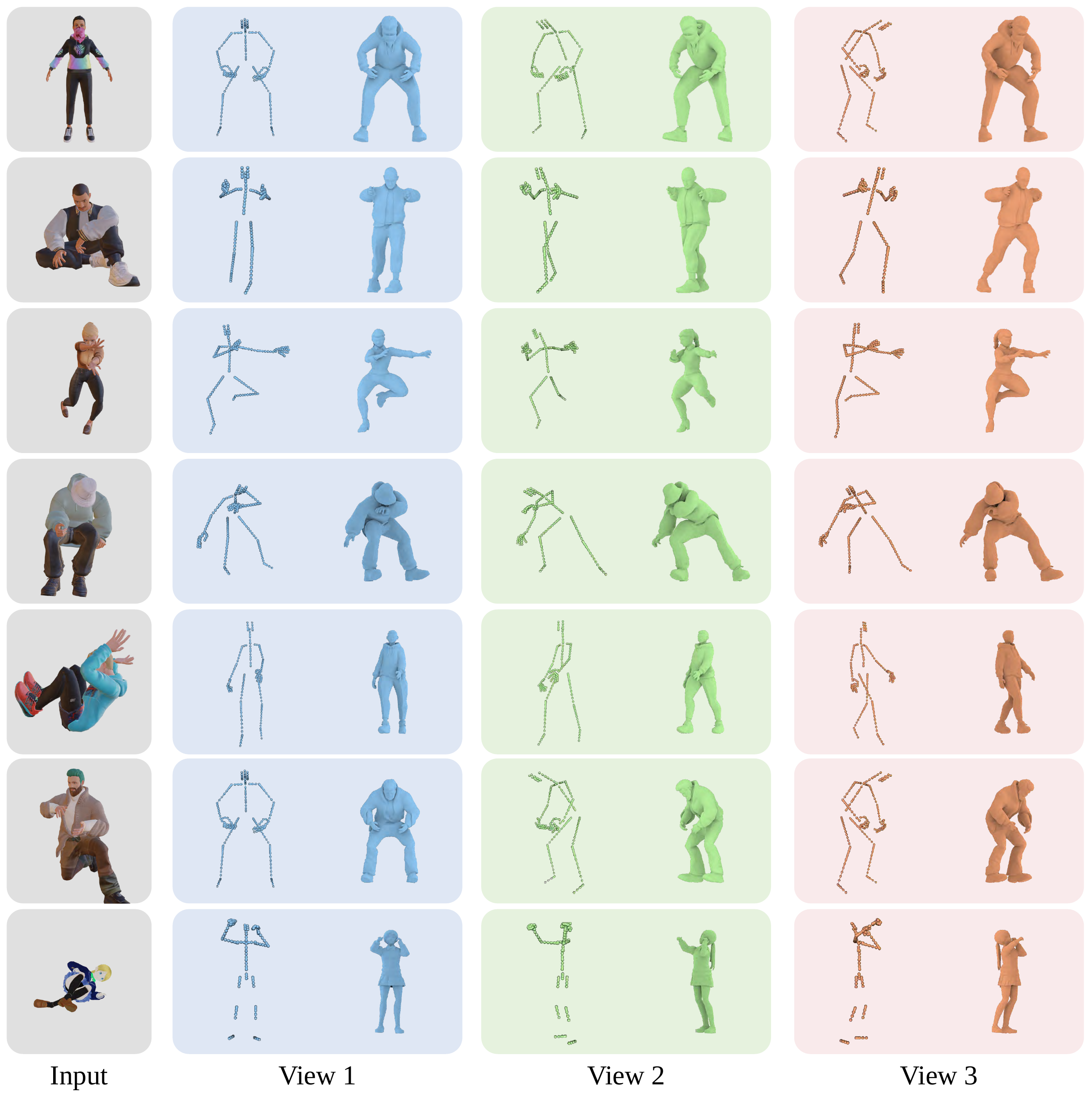}
    \centering
    \caption{The qualitative results for arbitrary-pose stylization.}
    \label{fig:arbitrary-pose-1}
\end{figure*}

\begin{figure*}[t]
\includegraphics[width=1\linewidth]{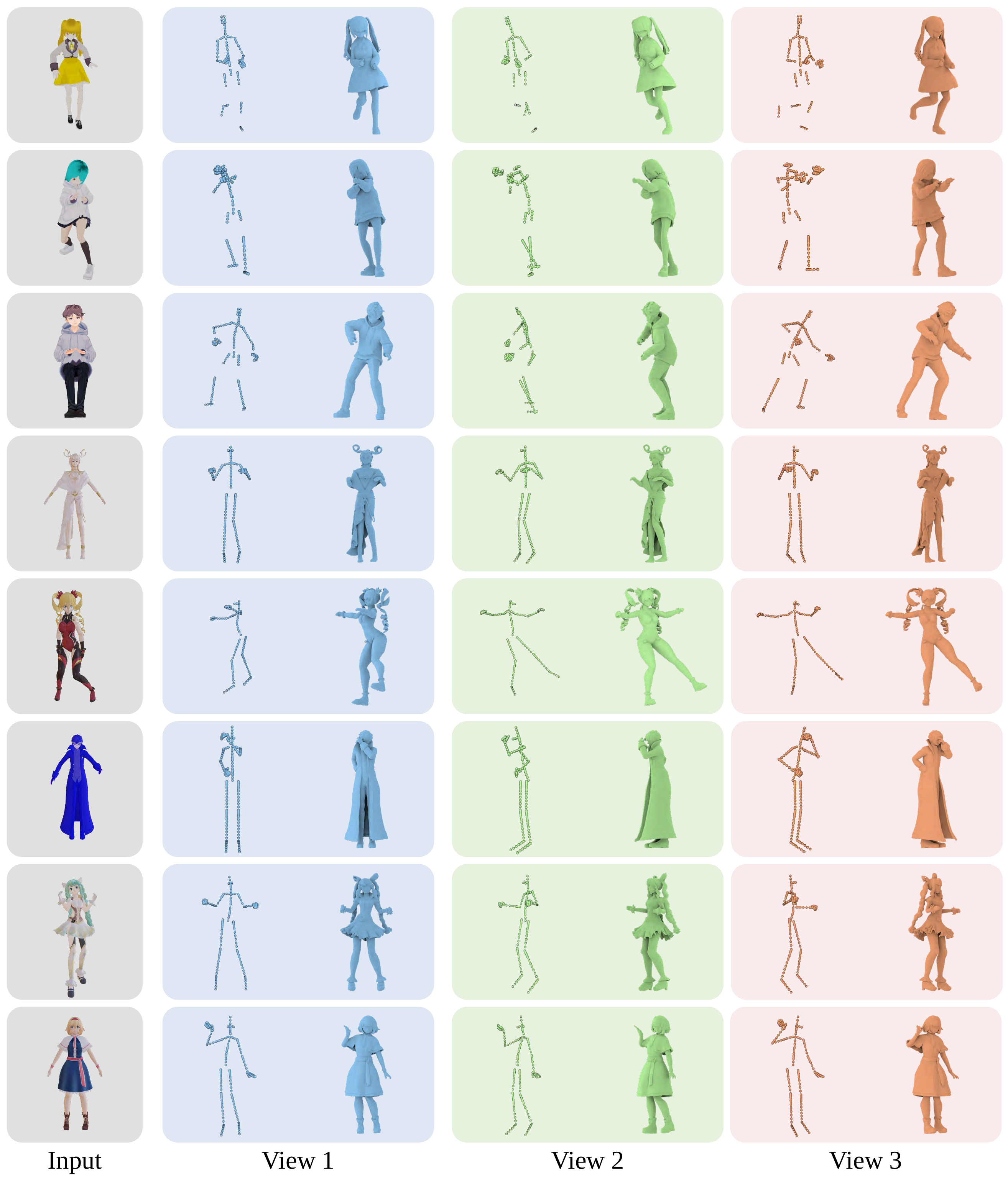}
    \centering
    \caption{The qualitative results for arbitrary-pose stylization.}
    \label{fig:arbitrary-pose-2}
\end{figure*}

Additional qualitative results for arbitrary-pose stylization are presented in Figure~\ref{fig:arbitrary-pose-1} and Figure~\ref{fig:arbitrary-pose-2}. These visualizations further demonstrate the robustness of PoseMaster. By seamlessly integrating pose stylization and 3D generation into a unified generative framework, we achieve precise pose control directly in the 3D domain. Moreover, the incorporation of 3D skeletons facilitates accurate pose recovery and manipulation, substantially enhancing the model's practical usability.

\subsection{Efficiency Analysis}

\begin{table}[t]
\caption{Time comparison of generation speed.}
\footnotesize
\setlength{\tabcolsep}{5.5pt}
\centering
\begin{tabular}{c|ccc}
\toprule
{Method} & CharacterGen~\cite{peng2024charactergen} &  StdGen~\cite{he2024stdgen} & Ours \\
\midrule
Inference time & $\sim$ 32.98 (s) & $\sim$ 61.54 (s) & $\sim$ 23.48 (s)   \\
\bottomrule
\end{tabular}
\label{tab:time}
\end{table}

We evaluate the computational efficiency of our method against previous multi-stage pose stylization approaches, including CharacterGen~\cite{peng2024charactergen} and StdGen~\cite{he2024stdgen}. All inference timings are measured on a single NVIDIA H20 GPU. As detailed in Table~\ref{tab:time}, PoseMaster achieves faster generation speeds. This acceleration is primarily attributed to our integrated, single-stage generative framework, which streamlines the conventional multi-stage workflow and eliminates intermediate computational bottlenecks.

\subsection{Robustness Analysis}
\begin{figure}[t]
\includegraphics[width=1\linewidth]{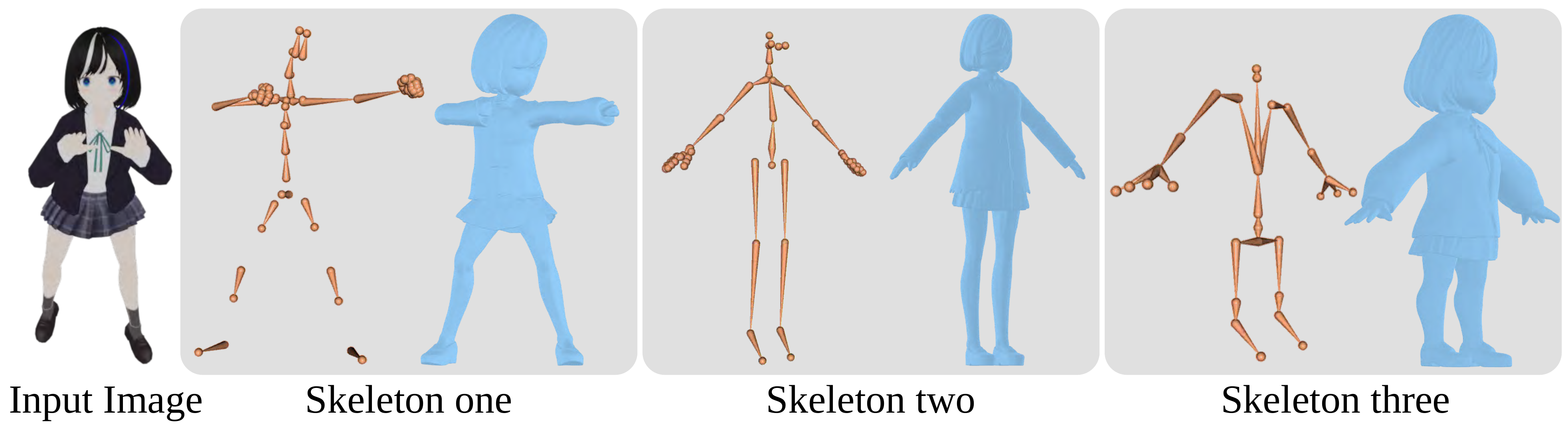}
    \centering
    \caption{The visualization for robustness analysis.}
    \label{fig:robust}
\end{figure}

\begin{figure*}[t]
\includegraphics[width=1\linewidth]{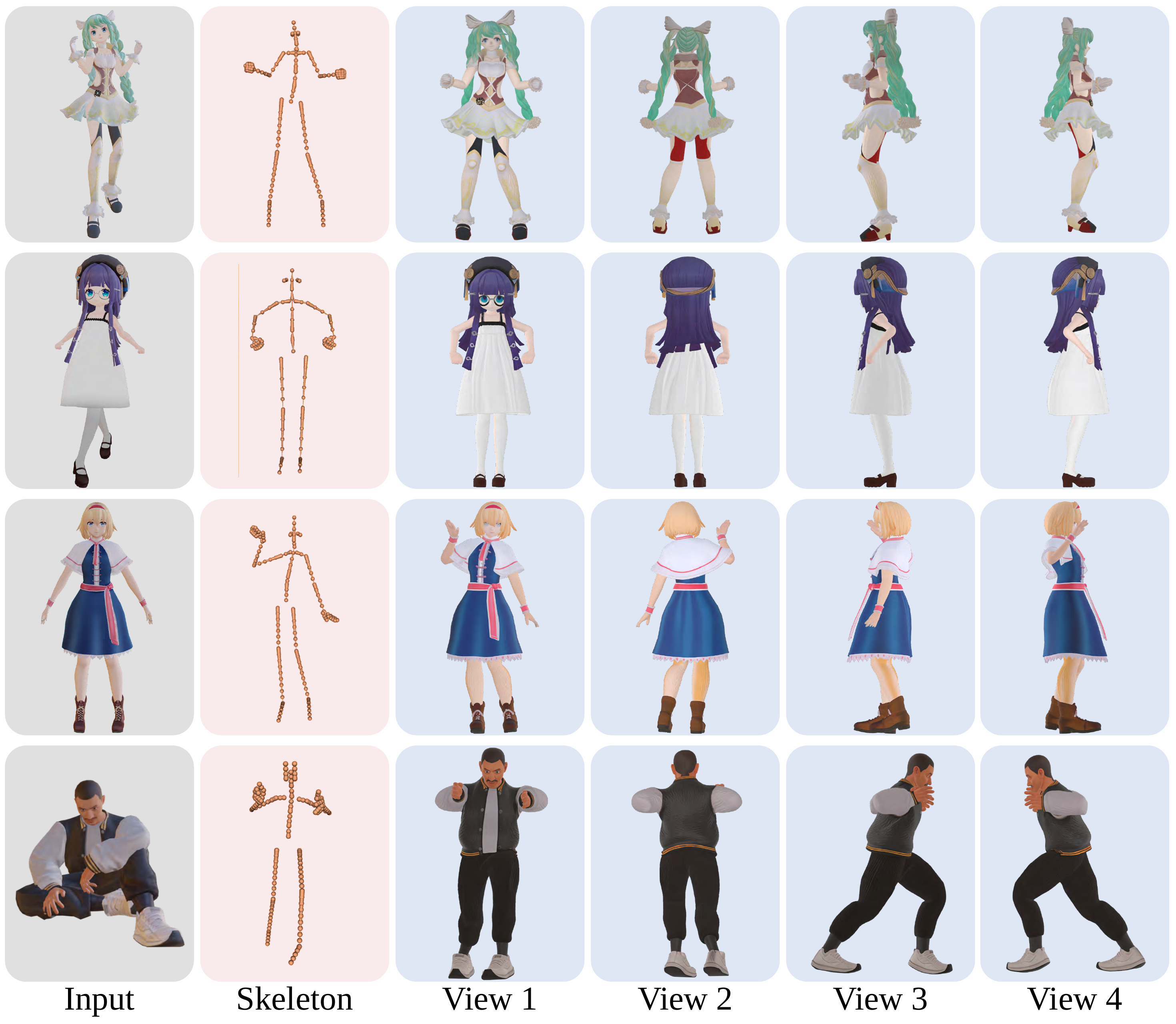}
    \centering
    \caption{The visualized results for textured mesh. We employ 3D native texturing approaches to synthesize textures for the meshes generated by PoseMaster. The final textured assets are rendered from four canonical viewpoints.}
    \label{fig:texture}
\end{figure*}

In this section, we introduce various skeleton conditions to validate the robustness of our model. As illustrated in Figure~\ref{fig:robust}, we design test cases featuring misaligned image-skeleton pairs with significant discrepancies in body proportions and topology. Remarkably, PoseMaster exhibits exceptional structural adherence. Even when conditioned on heavily mismatched skeletons, the model consistently synthesizes meshes that strictly conform to the target skeletal topology while faithfully preserving the appearance of the source image. This flexibility is particularly valuable for downstream game asset creation, such as character retargeting and body proportion customization

\subsection{Texture Generation}

In this work, geometry and texture synthesis are formulated as decoupled tasks; hence, explicit texture generation remains orthogonal to our main contributions. To demonstrate the practical utility of our approach, we provide texturing results leveraging state-of-the-art models. While the inherent pose inconsistency between the input image and the generated geometry poses a substantial challenge for texture mapping, we mitigate this by integrating advanced image-editing models (e.g., Qwen-Image~\cite{wu2025qwen}) and 3D native texturing approaches (e.g., UniTEX~\cite{liang2025unitex}, NaTex~\cite{lai2025natex}, LaFiTe~\cite{chen2025lafite}). As depicted in Figure.~\ref{fig:texture}, these frameworks demonstrate strong robustness against pose misalignment. For minor persistent artifacts, a promising future direction is to fine-tune these models specifically on misaligned image-geometry datasets.  Note that we employ Hunyuan3D 2.5~\cite{lai2025Hunyuan3D} as an intermediate geometric refinement step prior to the texturing phase.

\subsection{More Results on Pose Canonicalization}

{\bf Results on in-the-wild images.} 
To assess the zero-shot generalization capability of our model, we compile a diverse in-the-wild test set comprising both virtual avatars and real-world photographs. We evaluate our method against CharacterGen~\cite{peng2024charactergen}, StdGEN~\cite{he2024stdgen}, and Hunyuan3D 2.1~\cite{Hunyuan3D2025Hunyuan3D21}. For the Hunyuan3D 2.1, we utilized Qwen-Image~\cite{wu2025qwen} to transform the pose of the images to form its inputs. To circumvent the instability of Qwen-Image in generating A-pose outputs, we prompted the model to synthesize T-pose images instead. As shown in the qualitative comparisons, our method exhibits superior pose customization and geometric fidelity. CharacterGen and StdGEN, which are predominantly trained on single-style VRoid data, struggle to generalize to diverse artistic styles; their generated A-pose images often suffer from severe artifacts and distortions, fundamentally bottlenecking the subsequent 3D reconstruction. While Qwen-Image can alleviate some 2D distortion, controlling poses purely via text descriptions introduces significant randomness. For instance, in the inputs generated for Hunyuan3D, although the overall posture approaches a T-pose, critical details such as arm elevation, leg spacing, and perspective lack precise control. In contrast, our end-to-end framework bypasses the error-prone 2D editing stage. By employing explicit 3D skeletons as conditioning, we achieve strictly standardized pose canonicalization. Furthermore, inheriting the powerful priors of native 3D generation models ensures high-fidelity mesh topologies. 

\begin{figure*}[t]
\includegraphics[width=1\linewidth]{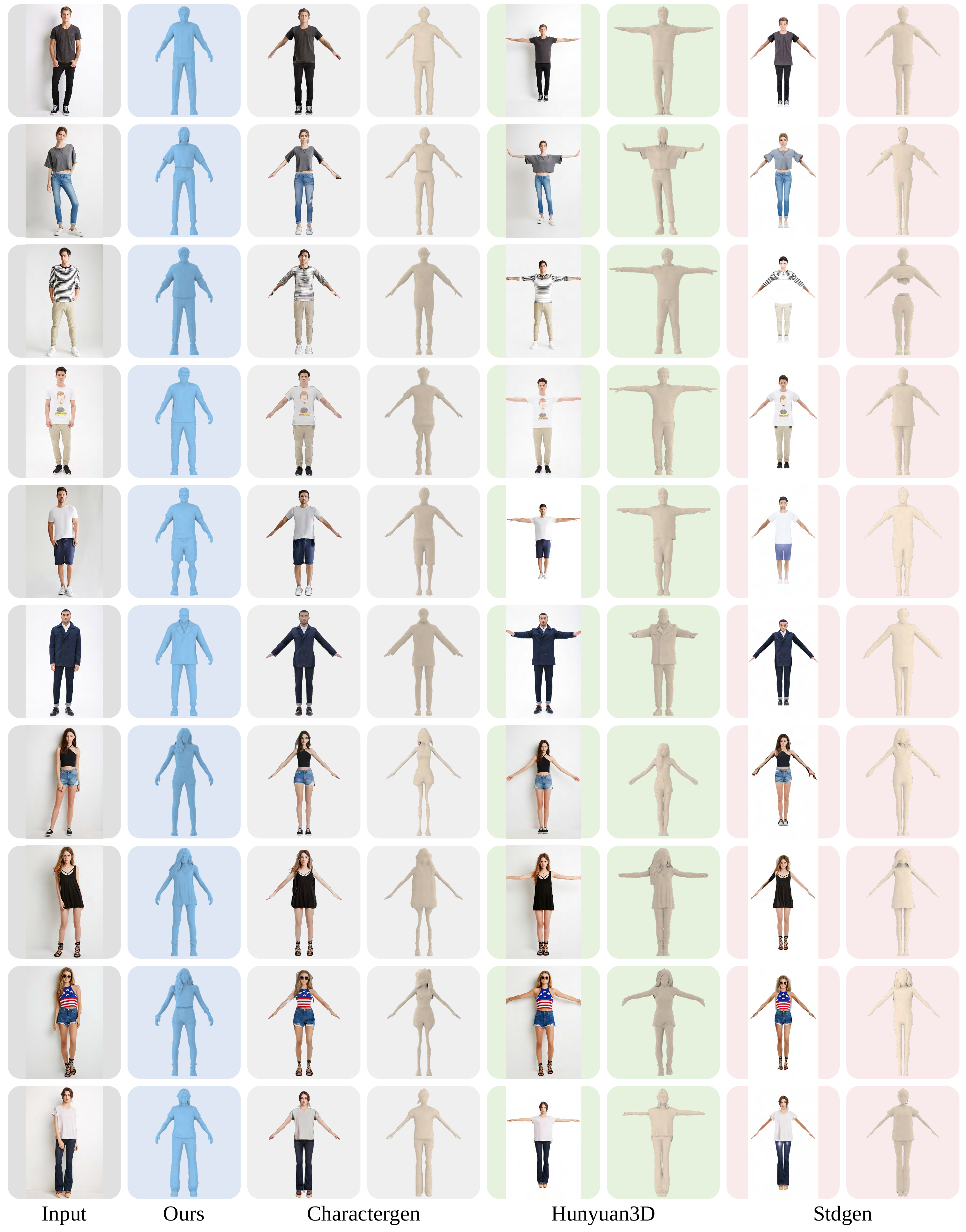}
    \centering
    \caption{The qualitative comparison for pose canonicalization on real-world images from the DeepFashion dataset~\cite{liuLQWTcvpr16DeepFashion}.}
    \label{fig:real_1}
\end{figure*}

Additionally, as illustrated in Figure~\ref{fig:real_1}, our method robustly handles real-world human images from the DeepFashion dataset~\cite{liuLQWTcvpr16DeepFashion}. The incorporation of a large-scale, diverse training dataset empowers our model to bridge the domain gap, significantly broadening its real-world applicability.

\begin{figure*}[t]
\includegraphics[width=1\linewidth]{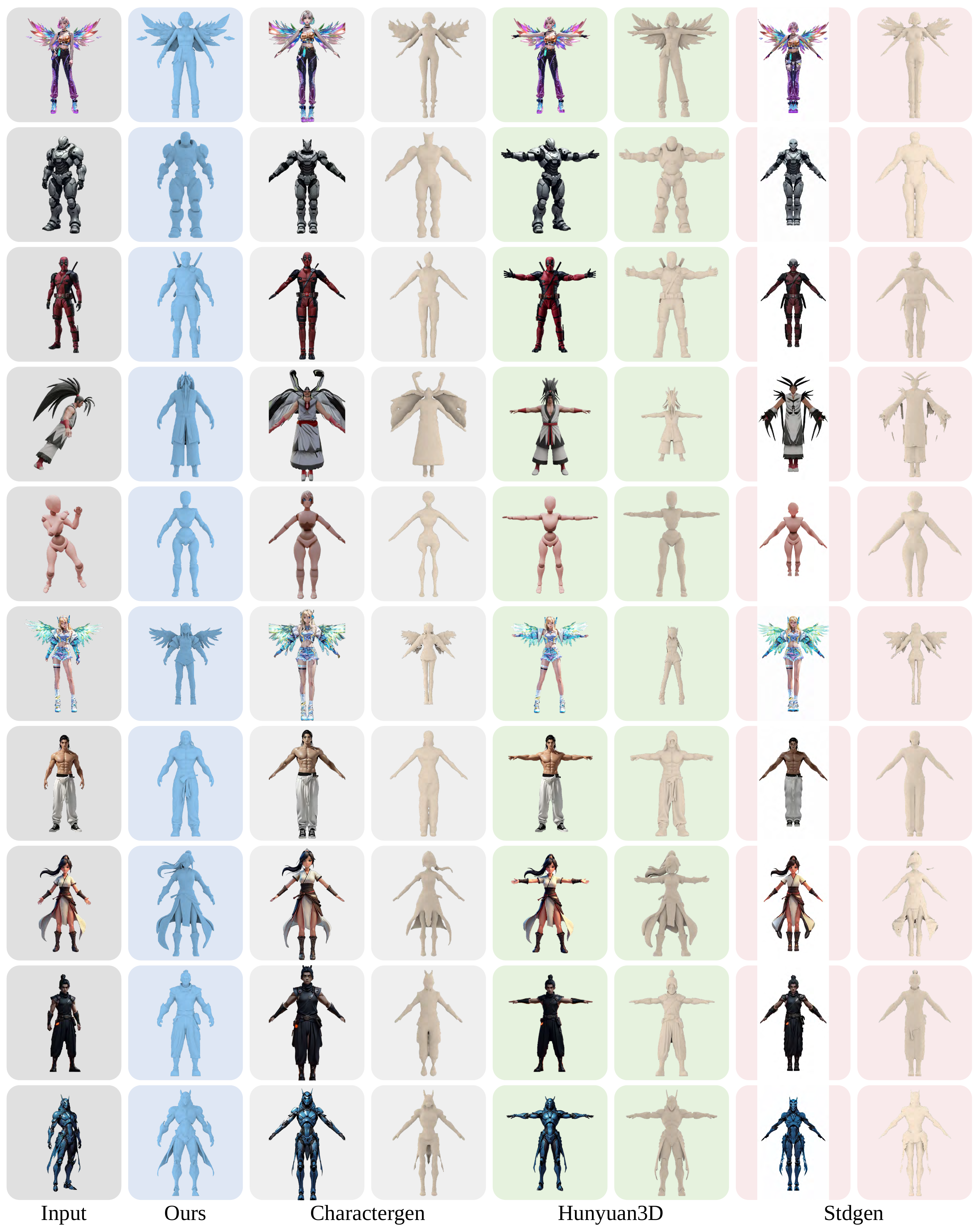}
    \centering
    \caption{The qualitative comparison for pose canonicalization on AI-synthesized images.}
    \label{fig:virtual}
\end{figure*}

{\bf Results on AI-synthesized images.} 
We further evaluate generalization on synthetic images produced by text-to-image models. As illustrated in Figure~\ref{fig:virtual}, PoseMaster exhibits superior stability in pose canonicalization tasks compared to pipeline-dependent approaches. Notably, for the Hunyuan3D baseline, the required T-pose inputs were synthesized via text-prompted Qwen-Image, which often struggles with complex pose transformations. 

{\bf Results on CharacterGen's testing set.} For a rigorous and fair comparison, we evaluate our method on the official test set provided by CharacterGen~\cite{peng2024charactergen}. As shown in Figure~\ref{fig:charactergen}, PoseMaster significantly outperforms both CharacterGen and StdGen~\cite{he2024stdgen} in terms of geometric quality and pose accuracy, further validating that native pose stylization yields optimal results for pose canonicalization.

\subsection{Application in 3D Printing}
\begin{figure}[t]
\includegraphics[width=1\linewidth]{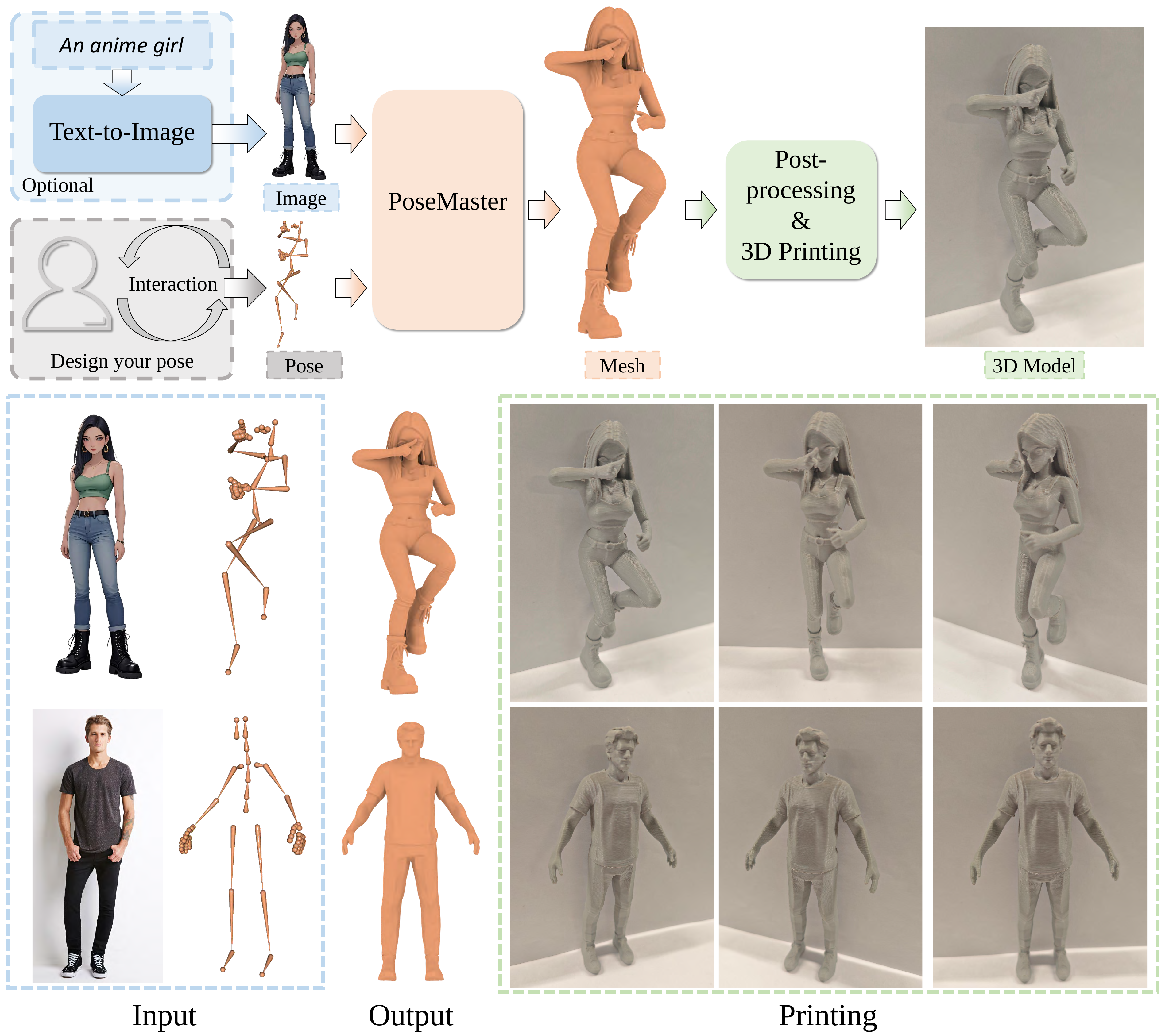}
    \centering
    \caption{ The system of 3D printing based on our PoseMaster. PoseMaster can customize the pose of a character from a single image for figure printing.}
    \label{fig:printing}
\end{figure}
Leveraging our arbitrary-pose stylization capabilities, we demonstrate a practical downstream application: a customized 3D printing pipeline for stylized anime characters. As depicted in Figure~\ref{fig:printing}, the workflow begins with a 2D anime character generated via text-to-image models. Users can then author desired poses using standard tools like Blender~\cite{blender} or Openpose Editor. By extracting the 3D skeleton from the authored pose, PoseMaster directly reconstructs a 3D character mesh matching both the input identity and the target pose. Following standard post-processing, the stylized mesh is readily 3D-printed.

\section{Limitation and Discussion}
First, we acknowledge that native 3D pose stylization remains a highly challenging task. While our framework demonstrates strong overall pose controllability, synthesizing fine-grained geometric details—such as intricate hand gestures, flowing skirts, and complex hairstyles—requires further exploration. Furthermore, our current implementation utilizes a single-stage generation paradigm operating at a spatial resolution of 512. Consequently, the high-frequency geometric fidelity of our outputs may lag behind recent state-of-the-art multi-stage refinement methods (e.g., Direct3D-S2~\cite{direct3ds2}, Hunyuan3D 2.5~\cite{lai2025Hunyuan3D}, and Sparc3D~\cite{li2025sparc3d}). A promising avenue for future work is to integrate a high-resolution geometric refinement module to upscale and enhance the coarse meshes generated in the first stage, thereby achieving photorealistic and highly detailed pose stylization.

\begin{figure*}[t]
\includegraphics[width=1\linewidth]{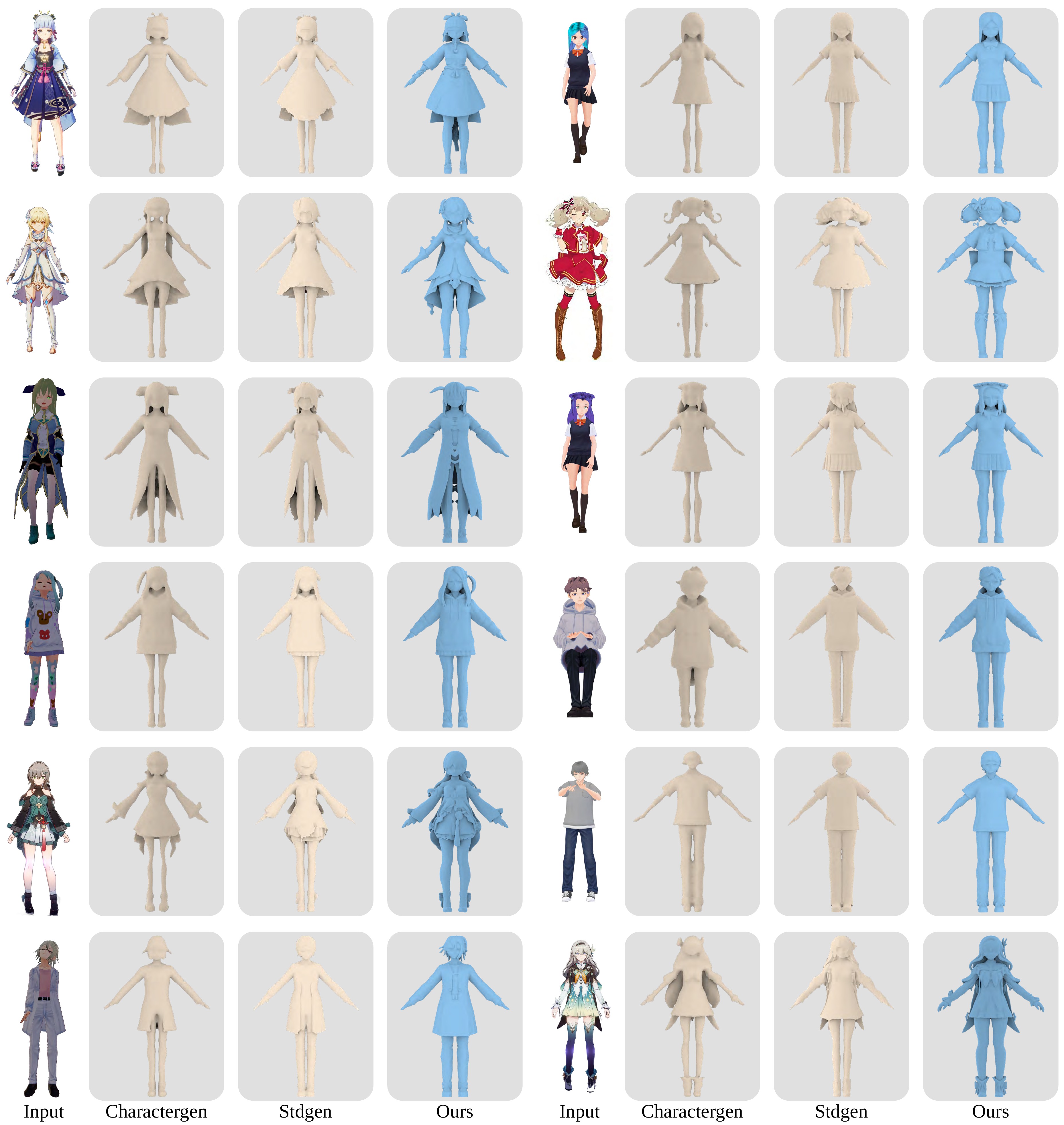}
    \centering
    \caption{The qualitative comparison for pose canonicalization on CharacterGen's~\cite{peng2024charactergen} testing dataset.}
    \label{fig:charactergen}
\end{figure*}

\end{document}